\documentclass[lettersize,journal]{IEEEtran}
\usepackage{amsmath,amssymb,amsfonts}
\usepackage{algorithmic}
\usepackage{algorithm}
\usepackage{array}
\usepackage{booktabs}
\usepackage{caption}
\usepackage{multirow}
\usepackage{subcaption}
\usepackage{textcomp}
\usepackage{stfloats}
\usepackage{url}
\usepackage{verbatim}
\usepackage{graphicx}
\usepackage{cite}
\usepackage[pagebackref,breaklinks,colorlinks]{hyperref}
\usepackage[capitalize]{cleveref}
\usepackage{diagbox}
\usepackage{xcolor}
\usepackage{footnote}
\usepackage{bbding}
\begin{document}

\title{High-Similarity-Pass Attention for Single Image Super-Resolution}

\author{Jian-Nan~Su,
	   Min~Gan\IEEEauthorrefmark{1},~\IEEEmembership{Senior Member,~IEEE,}
        Guang-Yong~Chen,
        Wenzhong~Guo,\\
        and~C. L. Philip Chen,~\IEEEmembership{Fellow,~IEEE}% <-this % stops a space
\IEEEcompsocitemizethanks{\IEEEcompsocthanksitem * Corresponding author (E-mail: aganmin@aliyun.com)\protect\\
%\IEEEcompsocthanksitem $\dagger$ https://github.com/laoyangui/SNLAN\protect\\
\IEEEcompsocthanksitem Jian-Nan~Su, Min~Gan, Guang-Yong~Chen and Wenzhong~Guo are with the College of Computer and Data Science, Fuzhou University, Fuzhou 350108, China.\protect\\
% Guo-Dong~Fan is with the College of Computer Science and Technology, Qingdao University, Qingdao, 266071, China
\IEEEcompsocthanksitem C. L. Philip Chen is with the School of Computer Science and Engineering, South China University of Technology, Guangzhou 510641, China, and also with the Faculty of Science and Technology, University of Macau, Macau, China
}}% <-this % stops an unwanted space
\maketitle

\begin{abstract}
Recent developments in the field of non-local attention (NLA) have led to a renewed interest in self-similarity-based single image super-resolution (SISR). Researchers usually used the NLA to explore non-local self-similarity (NSS) in SISR and achieve satisfactory reconstruction results. However, a surprising phenomenon that the reconstruction performance of the standard NLA is similar to the NLA with randomly selected regions stimulated our interest to revisit NLA. In this paper, we first analyzed the attention map of the standard NLA from different perspectives and discovered that the resulting probability distribution always has full support for every local feature, which implies a statistical waste of assigning values to irrelevant non-local features, especially for SISR which needs to model long-range dependence with a large number of redundant non-local features. Based on these findings, we introduced a concise yet effective soft thresholding operation to obtain high-similarity-pass attention (HSPA), which is beneficial for generating a more compact and interpretable distribution. Furthermore, we derived some key properties of the soft thresholding operation that enable training our HSPA in an end-to-end manner. The HSPA can be integrated into existing deep SISR models as an efficient general building block. In addition, to demonstrate the effectiveness of the HSPA, we constructed a deep high-similarity-pass attention network (HSPAN) by integrating a few HSPAs in a simple backbone. Extensive experimental results demonstrate that HSPAN outperforms state-of-the-art approaches on both quantitative and qualitative evaluations. %On the challenging test dataset Urban100, specially designed for testing self-similarity-based SISR methods, our HSPAN outperforms the most recent state-of-the-art method by 0.31dB, 0.38dB and 0.24dB at scale ×2, ×3, and ×4, respectively.%  of different degradation types (e.g. noise and blur) both qualitatively and quantitatively % Our code and a pre-trained model were uploaded to GitHub\IEEEauthorrefmark{2} for validation.
% The drawback is caused by the softmax transformation used in the NLA, which assigns non-zero weights to all non-local features.

\end{abstract}

\section{Introduction} \label{intro}
\IEEEPARstart{S}{ingle} Image Super-Resolution (SISR) is the process of recovering a high-resolution (HR) image from a corresponding low-resolution (LR) observation. SISR has been widely used for many important applications, including remote sensing, face recognition, and medical image processing \cite{lyu2020multi,cherukuri2019deep,li2019low}. However, SISR problems are often considered as ill-posed and heavily rely on efficient image priors. The high frequency of recurrence of small patches in a natural image provides powerful image-specific self-similarity priors\cite{glasner2009super} to regularize the SISR problems. These repeated internal textures are more closely related to the input LR image than external training datasets. Essentially, self-similarity offers an insightful solution based on similarity to explore non-local image textures,  which can filter out irrelevant non-local textures while discovering valuable ones. For instance, when repairing structured architectural textures, the associated non-local architectural regions are more meaningful than low-frequency face or background regions.
%The degradation process of the HR image with down-scaling factor $s$ is of the form:
%\begin{equation}
%\label{eq_hr_degradation}
%\boldsymbol{x}=\boldsymbol{d}(\boldsymbol{y};\boldsymbol{\alpha}).
%\end{equation}
%$\boldsymbol{y}\in R^{sh \times sw}$ and $\boldsymbol{x}\in R^{h \times w}$ are the HR image and the corresponding LR observation, respectively. $\boldsymbol{d}(\cdot)$ represents the downsampling operation with parameters $\boldsymbol{\alpha}$, as discussed in previous studies\cite{yang2010image,mei2021image}. More generally, we consider the degradation under additive noise and blurring conditions, which are closer to the real-world scenarios. The generalization of the degradation process to this non-ideal conditions is
%\begin{equation}
%\label{eq_hr_degradation_with_noise_blur}
%\boldsymbol{x}=\boldsymbol{d}(\boldsymbol{b}*\boldsymbol{y};\boldsymbol{\alpha}) + \boldsymbol{n},
%\end{equation}
%where $*$ is the convolution operation. $\boldsymbol{b}$ and $\boldsymbol{n}$ are blurring filter and additive noise, respectively.

\begin{figure}[t]
  \centering
  \includegraphics[width=\linewidth]{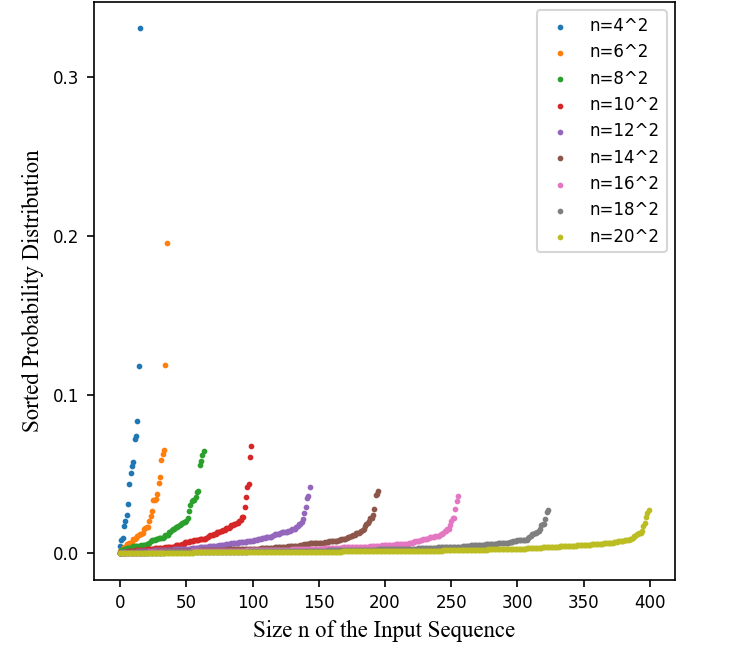}
   \caption{An illustration of the resulting probability distribution of the softmax transformation with different size of input sequences. The 9 input sequences were sampled from the standard normal distribution ${\mathcal {N}}(0 ,1)$.}
   \label{fig_similarity_limitations}
\end{figure}

In deep learning-based SISR models, self-similarity is usually explored through non-local attention (NLA)\cite{wang2018non}, which was originally used to model non-local dependencies for high-level vision tasks and has also been proven to be effective in SISR \cite{liu2018non,dai2019second,mei2020image,mei2021image}. These NLA-based SISR methods use the softmax transformation to assign attention weights to non-local information. However, as illustrated in \cref{fig_similarity_limitations}, we can see that as the size of non-local sequence increases, the resulting probability distribution of the softmax transformation is gradually flattened, i.e., \emph{valuable non-local information will be overwhelmed by a large amount of irrelevant non-local information}. Specifically, in short-range sequence modeling (e.g. $n=4^2$), the resulting probability distribution of the softmax transformation is distinguishable. Whereas in long-range sequence modeling (e.g. $n=20^2$), the resulting probability of most non-local information is close to zero, and even the most important non-local information can only be assigned to a very small value.

%The irrelevant non-local information can be regarded as noise signals, and the sum of the weights of these noise signals in the softmax transformation is much larger than the weight of valuable non-local signals. The reason for this drawback is that the softmax transformation cannot completely remove irrelevant information, resulting in suboptimal utilization of the non-local information.

\begin{figure}[t]
  \centering
  \includegraphics[width=0.9\linewidth]{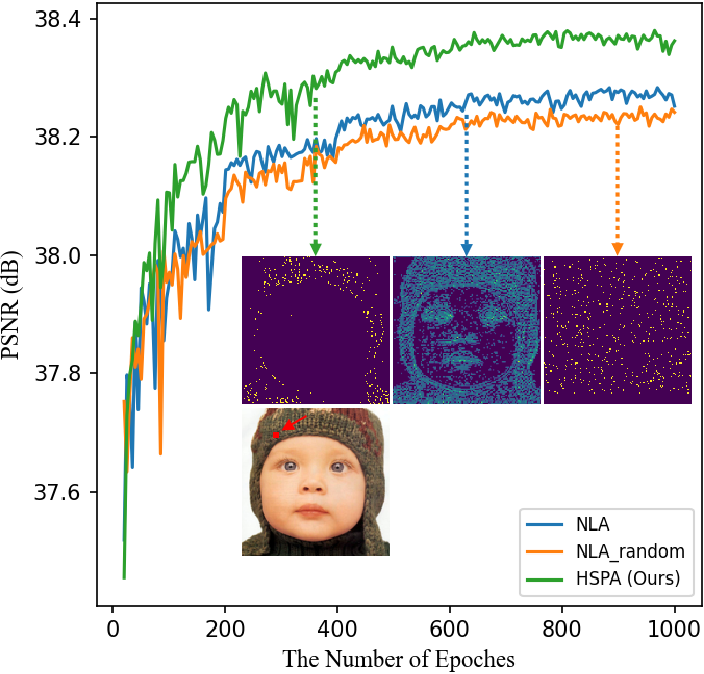}
   \caption{Comparisons between the NLA\cite{wang2018non} and our HSPA on Set5 ($\times 2$) during training. In NLA\_random, we randomly select 512 features for non-local fusion. Please zoom in for best view.}
   \label{fig_converge_curve_cmp}
\end{figure}

The above phenomenon shows that the softmax transformation, a key component of the standard NLA, may degrade the importance of valuable non-local information for long-range sequence modeling. 
Furthermore, it may introduce interference by incorporating irrelevant features, which can be regarded as noise, into the reconstruction result. Unfortunately, exploring the self-similarity in SISR involves long-range sequence modeling with a lot of irrelevant information. Therefore, we speculated that the softmax transformation would make the NLA inefficient in exploring the image self-similarity. To validate this speculation, we separately added the NLA and NLA\_random to our backbone network with 128 channels. In NLA\_random, we randomly select 512 feature vectors from the non-local feature space for non-local fusion, instead of using all non-local feature vectors like the NLA. Surprisingly, we found that the SR performance of NLA and NLA\_random is very close (see \cref{fig_converge_curve_cmp}). This motivated us to revisit the softmax transformation of the NLA for more effective exploration of self-similarity in deep SISR.

%From the perspective of image compression [xxx], which suggests that most of the information in nature images can be discarded with almost no perceptual loss. This also indicates that the non-local information should be modeled sparsely, instead of providing dense probability distribution like the softmax transformation. 

In many existing NLA-based SISR methods, the softmax transformation is commonly used to convert similarity vectors into probability distributions, such as NLRN\cite{liu2018non}, RNAN\cite{zhang2019residual}, SAN\cite{dai2019second}, and CSNLN\cite{mei2020image}. However, these methods require modeling the global features of long-range sequences, which leads to the issue of the softmax transformation mentioned above. Hence, NLSA\cite{mei2021image} decomposed the challenge of the long-range sequence modeling problem into a series of shorter sequence modeling sub-problems via locality sensitive hashing (LSH)\cite{gionis1999similarity}. Although NLSA avoided modeling long-range sequences, it may miss crucial long-range information due to the large variance of the LSH's estimation. To capture long-range information while alleviating the issue caused by the softmax transformation, ENLCA\cite{xia2022efficient} proposed to multiply similarity vectors by an amplifification factor, which enforces the non-local attention to give higher weights to related information. Unfortunately, this approach leads to an increase in approximation variance, making the selection of the amplifification factor non-robust.

In this paper, we focus on exploring a novel approach for weighting non-local information in deep SISR, providing a new perspective for existing NLA-based SISR methods. Consequently, we proposed the high-similarity-pass attention (HSPA) with a soft thresholding operation, which returns compact probability distributions by truncating small probability values (low-similarity) to zero. This characteristic enables our HSPA to remove irrelevant non-local information and identify a set of relevant information for image reconstruction, making deep SISR models more efficient and interpretable. From the attention maps in \cref{fig_converge_curve_cmp}, we can observe that our HSPA achieves superior SR performance than the NLA by fusing more related non-local self-similarity information. Crucially, we derived a closed-form Jacobian expression of the proposed soft thresholding operation to train our HSPA in an end-to-end manner. To the best of our knowledge, this is the first attempt to provide compact probability distributions with a closed-form Jacobian expression for backpropagation algorithms in self-similarity-based deep SISR models. With the proposed HSPA, we constructed a deep high-similarity-pass attention network (HSPAN) shown in \cref{fig_SNLAN_snlan}, from which we can see that our HSPAN is built on a  simple residual backbone with some HSPA-based modules (HSPAMs). Specifically, each HSPAM has a residual connection and consists of an HSPA, a locality exploration block (LEB) and a feature refinement convlution. The HSPA is responsible for capturing non-local information, while the LEB explores the locality inductive bias of nature images.

%The effectiveness of our SNLAN will be verified in the experiment section for different degradation types (e.g. noise and blur) of SISR tasks. In all degradation types, our SNLAN outperforms other state-of-the-art SISR models \cite{zhang2018image,dai2019second,niu2020single,mei2021image} by a large margin. In addition, we integrated our SNLA into some representative deep SISR models, such as FSRCNN \cite{dong2016accelerating}, EDSR\cite{lim2017enhanced}, and RCAN\cite{zhang2018image} to verify the universality of our SNLA in deep SISR. Experimental results show that our SNLA can significantly improve the reconstruction performance of deep SISR models from the simple FSRCNN\cite{dong2016accelerating} to the very complex RCAN\cite{zhang2018image}. Ablation studies are also conducted to analyze the effect of the proposed SNLA and LEB on the reconstruction performance. In summary, our experimental results show that the proposed SNLA is universal and efficient for SISR of different degradation types, and can be integrated as a general building block to explore the self-similarity in deep SISR models.

We believe our findings will stimulate further investigation and discussion about the use of NLA in self-similarity-based deep SISR models. Our contributions can be summarized as follows:
\begin{itemize}
\item{We provided new insights into the limitations of NLA used in self-similarity-based deep SISR methods and argued that the softmax transformation in NLA has insurmountable flaws for SISR with long-range sequences. (As shown in \cref{fig_similarity_limitations} and \cref{fig_converge_curve_cmp})}
\item{We formalized a concise yet effective soft thresholding operation and explored its key properties, which make it possible to optimize our high-similarity-pass attention (HSPA) end-to-end in deep SISR.}
\item{A deep high-similarity-pass attention network (HSPAN) was designed by using our HSPA-based modules (HSPAMs) and achieved state-of-the-art results both quantitatively and qualitatively.}
\end{itemize}

\section{Related work} \label{related_work}
Single image super-resolution (SISR) is a challenging problem, especially when the LR image has a limited number of pixels. To address this issue, various image priors have been proposed to stabilize the inversion of this ill-posed problem, such as the representative self-similarity\cite{glasner2009super,zontak2011internal} and sparse representation\cite{candes2006compressive,donoho2006compressed}. The proposed high-similarity-pass attention (HSPA) in this paper benefits from the above two priors, thus we limit our discussion here to SISR methods based on the two priors and classify the corresponding methods into two main categories: self-similarity-based SISR and sparse representation-based SISR.

\subsection{Self-similarity-based SISR}
There are many classical self-similarity-based SISR methods\cite{ebrahimi2007solving,protter2008generalizing,glasner2009super,freedman2011image,huang2015single} that have achieved satisfactory reconstruction results. The self-similarity prior provides a very efficient solution for SISR to explore non-local image information. The difference among these self-similarity-based SISR methods in utilizing self-similarity is mainly in the range of non-local search space. Yang \textit{et al.}\cite{yang2013fast} and Freedman \textit{et al.}\cite{freedman2011image} restricted the non-local search space to some specified local regions for reducing the patch search complexity. Ebrahimi \textit{et al.}\cite{ebrahimi2007solving} and Glasner \textit{et al.}\cite{glasner2009super} extended the non-local search space to cross-scale images to achieve more accurate SR performance. Huang \textit{et al.}\cite{huang2015single} further expanded the search space by modeling geometric transformations such as perspective distortion and local shape variation.

In deep learning-based SISR, the self-similarity is usually integrated by non-local attention (NLA)\cite{wang2018non}. Specifically, self-similarity prior is fused as a weighted sum over all pixel-wise features in the NLA. Liu \textit{et al.}\cite{liu2018non} proposed a non-local recurrent network as the first attempt to use the NLA in deep recurrent neural network for capturing long-range self-similarity in SISR and denoising. Then, Dai \textit{et al.}\cite{dai2019second} combined the NLA with channel-wise attention in deep convolutional neural network to simultaneously utilizing both spatial and channel features correlations for more powerful feature representation. Mei \textit{et al.}\cite{mei2020image} first attempted to integrate cross-scale self-similarity in deep learning-based SISR methods by the cross-scale non-local attention and achieved impressive SR performance. Luo \textit{et al.}\cite{luo2022multi} proposed a hash-learnable non-local attention to capture long-range self-similarity for lightweight SISR. Although these NLA-based deep SISR methods can generate impressive SR results, they all ignore the limitations (see \cref{fig_similarity_limitations}) of the softmax transformation used in the NLA, which leads to inefficient exploration of non-local self-similarity. This motivates us to explore a more efficient method to weight non-local self-similarity information for deep SISR models.
\subsection{Sparse representation-based SISR}
The sparse representation prior has been successfully integrated in many image processing tasks, such as SISR\cite{yang2010image,peleg2014statistical,wang2015deep,mei2021image} and denoising\cite{elad2006image,tian2020attention}. In SISR, the sparse representation suggests that HR images can be well-expressed by the sparse linear combinations of atoms in an appropriate over-complete dictionary\cite{yang2010image,yang2012coupled}. Yang \textit{et al.}\cite{yang2010image} proposed a joint dictionary learning method to make the sparse representation between LR and HR patch pairs consistent, thus the sparse representation of an LR image can be  directly used to generate the corresponding HR image patch with the HR dictionary. Kim \textit{et al.}\cite{kim2010single} constructed a sparse
basis set to reduce the time complexity of regression during both training and testing. Peleg \textit{et al.}\cite{peleg2014statistical} proposed a statistical prediction model based on sparse representation to predict the HR representation vector from its corresponding LR coefficients without using the invariance assumption of the sparse
representation. 

Dong \textit{et al.}\cite{dong2015image} first introduced the sparse
representation prior in a deep CNN-based SISR model with the ReLU activation enforced about 50\% sparsity by zeroing all negative features. Wang \textit{et al.}\cite{wang2015deep} combined the domain expertise of the sparse representation and deep learning models to achieve further improved SR results. Recently, Fan \textit{et al.}\cite{fan2020neural} explicitly integrated sparsity constraints in hidden neurons and showed that sparse representation is crucial in deep learning-based SISR. These sparse representation-based deep SISR methods only explored the sparse representation of deep spatial features, lacking the investigation of sparse non-local self-similarity. Mei \textit{et al.}\cite{mei2021image} and Su \textit{et al.}\cite{su2022global} tried to combine sparse representation with the exploration of non-local self-similarity in deep SISR models. Although they limited the scope of non-local exploration for sparsity, they ignored the essential reason why the NLA cannot explore non-local self-similarity sparsely. Inspired by these sparse representation-based deep SISR methods and our analysis of the limitations of the softmax transformation in NLA, we integrated sparse representation into non-local self-similarity exploration by incorporating a soft thresholding operation into the high-similarity-pass attention.

\begin{figure*}[!htbp]
  \centering
  \includegraphics[width=\linewidth]{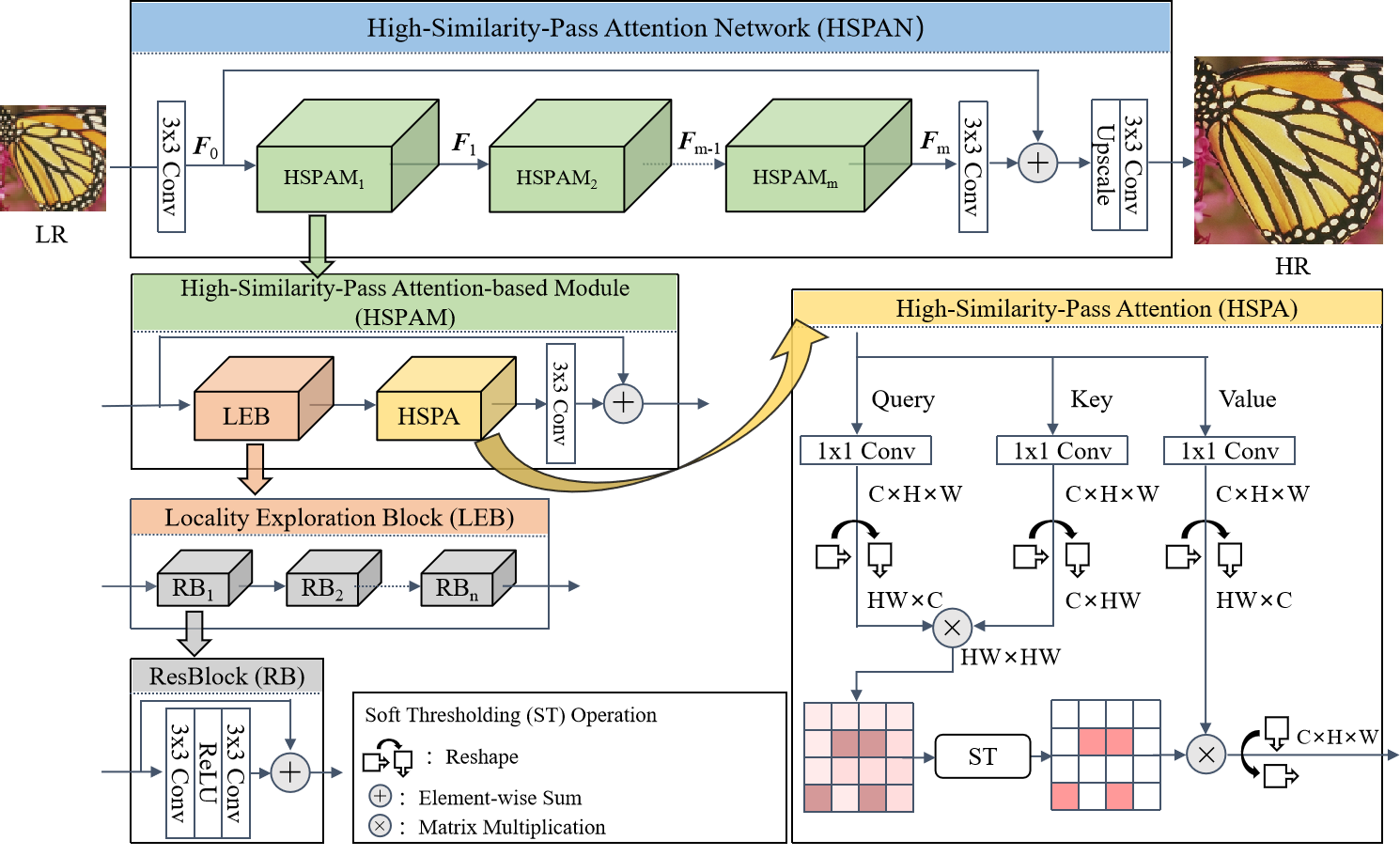}
   \caption{An illustration of our HSPAN. The implementation details of the ST operation can be found in \cref{alg:st}.}% For the convenience of description, we only show the process of i-th bucket in Global Learnable Attention (GLA) and the processes of other parallel execution buckets are the same as those of the i-th bucket. Furthermore, the implementation details of the process of the i-th bucket in GLA is described in \cref{alg:gla}.
   \label{fig_SNLAN_snlan}
\end{figure*}

\section{Methodology}
In this section, we will introduce the details of our deep high-similarity-pass attention network (HSPAN). The HSPAN consists of a simple residual backbone\cite{zhang2018image,su2022global} and some high-similarity-pass attention-based modules (HSPAMs). We start with an overview of the HSPAN and then introduce the details of the HSPAM.

\subsection{An Overview of the HSPAN} \label{snlan}
As illustrated in \cref{fig_SNLAN_snlan}, our HSPAN is an end-to-end SISR network consisting of three parts: LR features extraction, local and non-local deep features fusion, and HR reconstruction. In  LR features extraction part, the shallow features $\boldsymbol{F}_0$ are extracted from the given LR image $\boldsymbol{x}$ by a convolutional layer with trainable parameters $\boldsymbol{\theta}$. This process can be formulated as
\begin{equation}
\boldsymbol{F}_0=\phi(\boldsymbol{x};\boldsymbol{\theta}),
  \label{eq_shallow_feature_extract}
\end{equation}
where $\phi(\cdot)$ is a convolution operation. Then, $\boldsymbol{F}_0$ is fed into the local and non-local deep features fusion part with $m$ HSPAMs
\begin{equation}
\boldsymbol{F}_m=\varphi(\boldsymbol{F}_0;\boldsymbol{\delta}),
  \label{eq_deep_feature_extract}
\end{equation}
where $\varphi(\cdot)$ is the function of our local and non-local deep features fusion part with trainable parameters $\boldsymbol{\delta}$ and $\boldsymbol{F}_m$ are the output features. Finally, the obtained deep features $\boldsymbol{F}_m$ are upscaled by sub-pixel operation $\uparrow$ \cite{shi2016real} and then used to reconstruct an HR image $\boldsymbol{\hat{\boldsymbol{y}}}$ in the HR reconstruction part
\begin{equation}
\boldsymbol{\hat{\boldsymbol{y}}}=\psi(\boldsymbol{F}_m\uparrow;\boldsymbol{\alpha}),
  \label{eq_hr_image_reconstruction}
\end{equation}
where $\psi(\cdot)$ represents the HR reconstruction part with trainable parameters $\boldsymbol{\alpha}$ for the final RGB image reconstruction. The above three parts can be simplified as follow
\begin{equation}
\boldsymbol{\hat{\boldsymbol{y}}}=\text{HSPAN}(\boldsymbol{x};(\boldsymbol{\theta},\boldsymbol{\delta},\boldsymbol{\alpha})),
  \label{eq_snlan}
\end{equation}
where $\text{HSPAN}(\cdot)$ is the function of our deep sparse non-local attention network. As suggested in previous works\cite{kim2016accurate,zhang2018image}, we use a long residual connection to directly bypassed abundant low-frequency information and prevent gradients from exploding.

%The mean absolute error (MAE) is used as the loss function to train our DLSN. Let us consider a set of LR images denoted by $\{\boldsymbol{I}_{i}\}$ and the corresponding HR images denoted by $\{\boldsymbol{H}_{i}\}$, then the MAE loss function $L(\cdot)$ is given by:
%\begin{equation}
%L(\boldsymbol{I}_{i};\boldsymbol{\alpha},\boldsymbol{\beta},\boldsymbol{\gamma},\uparrow) = \frac{1}{N}\sum_{i=1}^{N}||DLSN(\boldsymbol{I}_i;(\boldsymbol{\alpha},\boldsymbol{\beta},\boldsymbol{\gamma},\uparrow))-\boldsymbol{H}_i||_1,
%\label{eq_l1_loss}
%\end{equation}
%where $N$ is the number of training samples. More details of training would be shown in our experiments. Global learnable attention-based feature fusion module
\subsection{High-Similarity-Pass Attention-based Module (HSPAM)} \label{snlaffm}
The structure of our HSPAM is shown in \cref{fig_SNLAN_snlan}, from which we can see that the HSPAM serves as a basic block of our HSPAN. Specifically, each HSPAM is also with a residual connection and consists of a high-similarity-pass attention (HSPA), a locality exploration block (LEB) and a feature refinement layer. The HSPA is responsible for exploring non-local self-similarity information and the LEB is used to capture locality inductive bias.

The function $\Omega_i(\cdot)$ of the \emph{i}-th HSPAM can be formulated as follows:
\begin{equation}
\boldsymbol{F}_i=\Omega_i(\boldsymbol{F}_{i-1})=\Omega_i(\Omega_{i-1}(\cdots \Omega_2(\Omega_1(\boldsymbol{F}_0)))),
\label{eq_snlaffm}
\end{equation}
where $\boldsymbol{F}_{i}$ and $\boldsymbol{F}_{i-1}$ are the output and input of the \emph{i}-th HSPAM. The trainable parameters in HSPAM are omitted for simplicity.

\subsubsection{Locality Exploration Block (LEB)} \label{leb}
\ \par
We proposed the LEB to utilize the locality of the convolution layer for capturing the locality inductive bias of nature images, which plays a crucial role in image restoration problems. The structure details of the LEB are shown in \cref{fig_SNLAN_snlan}, from which we can see that the LEB is constructed by $n$ simplified convolution residual blocks. 

\subsubsection{High-Similarity-Pass Attention (HSPA)} \label{snla}
As discussed in \cref{intro}, the softmax transformation greatly affects the performance of NLA for modeling long-range sequence with a lot of irrelevant information. To solve the drawback caused by the softmax transformation, we proposed the HSPA (see \cref{fig_SNLAN_snlan}) with a soft thresholding (ST) operation that can effectively model long-range sequence with a lot of irrelevant information by generating sparse probability distributions. 

NLA is a pixel-level attention mechanism used in deep SISR methods \cite{zhang2019residual,dai2019second}, it calculates the influence of each feature vector on the query feature vector. The input feature maps $\boldsymbol{X}\in R^{H\times W\times C}$ of NLA are first reshaped into $\boldsymbol{X^{'}}\in R^{HW\times C}$ for illustration. Then, the response of the query feature vector $\boldsymbol{x}_i$ in NLA can be defined as 
\begin{equation}
  \text{NLA}(\boldsymbol{x}_i)=\sum_{j=1}^{N}\frac{{\rm exp}(d(\boldsymbol{x}_i,\boldsymbol{x}_j))}{\sum_{k=1}^{N}{\rm exp}(d(\boldsymbol{x}_i,\boldsymbol{x}_k))}\phi_v(\boldsymbol{x}_j),
  \label{eq_nla}
\end{equation}
where $\boldsymbol{x}_j$ and $\boldsymbol{x}_k$ are the \emph{j}-th and \emph{k}-th feature vectors on $\boldsymbol{X^{'}}$ respectively and $N=HW$. $\phi_v(\cdot)$ is a value vector generation function with $1\times 1$ convolution operation. $d(\cdot , \cdot )$ is used to measure the dot product similarity between feature vectors and can be expressed as
\begin{equation}
d(\boldsymbol{x}_i,\boldsymbol{x}_j)=\phi_q(\boldsymbol{x}_i)^\mathrm{T}\phi_k(\boldsymbol{x}_j),
  \label{eq_dot_product}
\end{equation}
where $\phi_q(\cdot)$ and $\phi_k(\cdot)$ are introduced to generate query and key feature vectors. From Eq. \eqref{eq_nla} and Eq. \eqref{eq_dot_product}, we can see that NLA provides a non-local information fusion scheme based on dot product similarity, which assigns more weights to feature vectors with higher similarity. Furthermore, the reason why the NLA cannot generate sparse probability distribution is that the numerator of the softmax transformation is always greater than zero, which allows two completely unrelated or even negatively correlated feature vectors to be given a positive weight.

In our HSPA, with Eq. \eqref{eq_dot_product}, we can obtain the similarity vector $\boldsymbol{s}$ of the query feature vector $\boldsymbol{x}_i$. Then, the weight of \emph{j}-th non-local feature vector is assigned by the proposed soft thresholding (ST) operation. Finally, the response of the query feature vector $\boldsymbol{x}_i$ can be formulated as
\begin{equation}
  \text{HSPA}(\boldsymbol{x}_i)=\sum_{j=1}^{N}\text{ST}_{j}(\boldsymbol{s})\phi_v(\boldsymbol{x}_j),
  \label{eq_snla}
\end{equation}
where $\text{ST}_{j}(\boldsymbol{s})$ is the \emph{j}-th component of the results of the ST operation and defined as
\begin{equation}
  \text{ST}_{j}(\boldsymbol{s})=\text{max}\{s_j-\kappa(\boldsymbol{s}),0\},
  \label{eq_st}
\end{equation}
where $s_j$ is the \emph{j}-th component of $\boldsymbol{s}$ and $\kappa(\cdot):\mathbb{R}^N\rightarrow\mathbb{R}$ is a soft threshold function that satisfies 
\begin{equation}
  \sum_{j=1}^{N}\text{max}\{s_j-\kappa(\boldsymbol{s}),0\}=1
  \label{eq_threshold_k_constrain}
\end{equation}
To derive the soft threshold function $\kappa(\cdot)$, we first sorted the similarity vector $\boldsymbol{s}$ and let the sorted $\boldsymbol{s}$ satisfy $s_{(1)}\geq s_{(2)}\geq \cdot \cdot \cdot \geq s_{(N)}$. Let $K=\{1,...,N\}$ and define $\mathcal{T}$ as
\begin{equation}
  \mathcal{T} =\text{max}\{k\in K\ |\ ks_{(k)} + 1 > \sum_{j=1}^{k}s_{(j)}\}.
  \label{eq_threshold_t}
\end{equation}
Then, the soft threshold function $\kappa(\cdot)$ can be derived from Eq. \eqref{eq_threshold_k_constrain} and Eq. \eqref{eq_threshold_t} as follows 
\begin{equation}
\begin{split}
  (\sum_{j=1}^{\mathcal{T}}s_{(j)})-\kappa(\boldsymbol{s})\times \mathcal{T}=1,\\
  \kappa(\boldsymbol{s})=\frac{(\sum_{j=1}^{\mathcal{T}}s_{(j)})-1}{\mathcal{T}}.
  \label{eq_threshold_k}
\end{split}
\end{equation}

%\begin{figure}[t]
%  \centering
%  \includegraphics[width=0.65\linewidth,height=0.55\linewidth]{images/snla.png}
%   \caption{An illustration of our SNLA. The implementation details of the ST operation can be found in \cref{alg:st}.}
%   \label{fig_SNLAN_snlan}
%\end{figure}

The details of the ST operation are shown in \cref{alg:st}, from which we can see that the core of the ST operation is to calculate the result of the soft threshold function $\kappa(\cdot)$. Then, all coordinates below this threshold will be truncated to zero, and the others will be shifted by this threshold. Obviously, our ST operation is not only able to generate a sparse probability distribution, but also preserves the basic properties of the softmax transformation: all probability values are greater than zero and sum to 1, and it assigns more weights to feature vectors with higher similarity.

\subsection{Properties of the ST operation} \label{property_st}
From optimization viewpoint, the attention weight vector $\tilde{\boldsymbol{s}}$ obtained by the ST operation can be regarded as the projection of the original similarity vector $\boldsymbol{s}$ onto the simplex $\Delta=\{\boldsymbol{p}\in \mathbb{R}^{N}|\sum{p_i}=1,p_i\geq0\}$\cite{duchi2008efficient,martins2016softmax}. That is:
\begin{equation}
\tilde{\boldsymbol{s}}=\underset{\boldsymbol{p}\in \Delta}{\text{argmin}}||\boldsymbol{p}-\boldsymbol{s}||^2.
  \label{eq_simplex}
\end{equation}
In SISR, exploring for non-local information is a long-sequence modeling problem, where the sequence length $N$ is usually greater than ten thousand. This makes the projection tend to hit the boundary of the simplex, leading to a sparse probability distribution. 
\begin{algorithm}[t]
\caption{Soft Thresholding (ST) Operation.}\label{alg:st}
\begin{algorithmic}[1]
\STATE \textbf{Input:} the similarity vector $\boldsymbol{s}$.
\STATE $\boldsymbol{s^{'}}\gets \boldsymbol{s}$.
\STATE $\boldsymbol{s} \gets {\rm {Sort}}(\boldsymbol{s})$
\ \textcolor{gray}{\# Let $s_{(1)}\geq s_{(2)}\geq \cdot \cdot \cdot \geq s_{(N)}$.}
\STATE $\mathcal{T}={\rm {max}}\{k\in K\ |\ ks_{(k)} + 1 > \sum_{j=1}^{k}s_{(j)}\}$
\STATE $\kappa(\boldsymbol{s})=\frac{(\sum_{j=1}^{\mathcal{T}}s_{(j)})-1}{\mathcal{T}}$
\STATE \textbf{Output:} $\text{max}\{\boldsymbol{s^{'}}-\kappa(\boldsymbol{s}),0\}$
\end{algorithmic}
\end{algorithm}

From the perspective of probability theory, we assume the similarity scores $s_1, s_2,..., s_N$ as random variables, and the corresponding order statistic are denoted as $s_{(1)}\geq s_{(2)}\geq \cdot \cdot \cdot \geq s_{(N)}$. Event $[\mathcal{T}>k]$ refers to the number of non-zero elements in the results of the ST operation is greater than $k$, which implies $(k+1)s_{(k+1)} + 1 > \sum_{j=1}^{k+1}s_{(j)}$. Thus, the probability of the event $[\mathcal{T}>k]$ can be estimated as
\begin{equation}
\begin{aligned}
  P([\mathcal{T}>k]) &= P([(k+1)s_{(k+1)} + 1 > \sum_{j=1}^{k+1}s_{(j)}])\\
  &= P([(s_{(1)}-s_{(k+1)})+ ... + (s_{(k)}-s_{(k+1)})] < 1)\\
  &\leq P([s_{(k)}-s_{(k+1)}<\frac{1}{k}]).
  \label{eq_event_probability}
\end{aligned}
\end{equation}
With the increase of $k$, the probability that the number of non-zero elements in the attention weight vector is greater than $k$ will become small, which is consistent with our intuitive motivation for designing the ST operation. 

In practice, the ST operation will cause many non-local weights to be truncated to zero, which means that it is not necessary to use all elements in the similarity vector $\boldsymbol{s}$. Therefore, we can use top-k instead of the sorting algorithm to reduce the computational complexity. The effects of different $k$ on reconstruction results will be investigated in the experiment section.

To train the HSPAN in an end-to-end manner, it is necessary to derive the closed-form Jacobian matrix of the proposed ST operation for backpropagation algorithms. From Eq. \eqref{eq_st}, we have
\begin{equation}
  \frac{\partial \text{ST}_{j}(\boldsymbol{s})}{\partial s_k}=\left\{\begin{matrix}
\delta_{\emph{jk}}-\frac{\partial \kappa(\boldsymbol{s})}{\partial s_k},\ \text{if}\ s_j>\kappa(\boldsymbol{s}),
\\ 
0,\ \ \ \ \ \ \ \ \ \ \ \ \text{if}\ s_j \leq \kappa(\boldsymbol{s}),
\end{matrix}\right.
  \label{eq_sto_gradient}
\end{equation}
where $\delta_{\emph{jk}}$ is the Kronecker delta which returns 1 if the variables $i$ and $j$ are equal, and 0 otherwise. The gradient of the soft threshold function $\kappa(\cdot)$ can be expressed as
\begin{equation}
  \frac{\partial \kappa(\boldsymbol{s})}{\partial s_k}=\left\{\begin{matrix}
\frac{1}{\mathcal{T}},\ \text{if}\ k \in \text{T}(\boldsymbol{s}),
\\ 
0,\ \ \ \ \text{if}\ k \notin \text{T}(\boldsymbol{s}),
\end{matrix}\right.
  \label{eq_threshold_func_gradient}
\end{equation}
where $\text{T}(\boldsymbol{s})=\{j\in K\ |\ \text{ST}_{j}(\boldsymbol{s})>0\}$ is the support set of the ST operation, $\mathcal{T}$ is the number of elements in $\text{T}(\boldsymbol{s})$, and $\boldsymbol{c}$ is the characteristic vector whose \emph{j}-th component is 1 if $j\in\text{T}(\boldsymbol{s})$, and 0 otherwise. We can combine Eq. \eqref{eq_sto_gradient} and Eq. \eqref{eq_threshold_func_gradient} to obtain the derivatives of the ST operation with respect to the variable $s_k$
\begin{equation}
  \frac{\partial \text{ST}_{j}(\boldsymbol{s})}{\partial s_k}=
\delta_{\emph{jk}}-\frac{c_{k}c_{j}}{\mathcal{T}}.
  \label{eq_sto_derivative_combine}
\end{equation}
%Because $k \leq t(\boldsymbol{s})$ and $s_j>\kappa(\boldsymbol{s})$ correspond to the same constraint space, we can combine Eq. \eqref{eq_sto_gradient} and Eq. \eqref{eq_threshold_func_gradient} into the following form
%\begin{equation}
%  \frac{\partial \text{ST}_{j}(\boldsymbol{s})}{\partial s_k}=\left\{\begin{matrix}
%\delta_{\emph{jk}}-\frac{1}{t(\boldsymbol{s})},\ \text{if}\ j,k \leq t(\boldsymbol{s}),
%\\ 
%0,\ \ \ \ \ \ \text{otherwise}.
%\end{matrix}\right.
%  \label{eq_sto_gradient_com}
%\end{equation}
%Let $\text{T}(\boldsymbol{s})=\{j\in K\ |\ \text{ST}_{j}(\boldsymbol{s})>0\}$ be the support set of the ST operation and $\boldsymbol{c}$ be the characteristic vector whose \emph{j}-th component is 1 if $j\in\text{T}(\boldsymbol{s})$, and 0 otherwise.
Then, the Jacobian matrix of the ST operator can be formulated as \begin{equation}
\boldsymbol{J}(\boldsymbol{s})=\text{Diag}(\boldsymbol{c})-\frac{\boldsymbol{c}\boldsymbol{c}^T}{\mathcal{T}},
  \label{eq_sto_derivative_combine_final}
\end{equation}
where $\text{Diag}(\boldsymbol{c})$ is a matrix with $\boldsymbol{c}$ in the diagonal.
%$|\text{T}(\boldsymbol{s})|$ is the number of elements in $\text{T}(\boldsymbol{s})$
%The ST operation is differentiable everywhere except at some splitting points where the support set $\text{T}(\boldsymbol{s})$ changes such as $\text{T}(\boldsymbol{s})\neq \text{T}(\boldsymbol{s}+\boldsymbol{\varepsilon})$ for some infinitesimal $\boldsymbol{\varepsilon}$.

From Eq. \eqref{eq_sto_derivative_combine_final} we can observe the special structure presenting in the Jacobian matrix that is obtained by subtracting a rank 1 matrix from a diagonal matrix. The special structure allows efficient calculation of the gradient. Given the feedback error $\boldsymbol{r}$, the gradient of the ST operation with respect to $\boldsymbol{s}$ can be expressed as
\begin{equation}
\bigtriangledown_{s}\text{ST}(\boldsymbol{s})=\boldsymbol{J}(\boldsymbol{s})\boldsymbol{r}=\boldsymbol{c}\odot(\boldsymbol{r}-\frac{\boldsymbol{c}^T\boldsymbol{r}}{\mathcal{T}}\cdot\boldsymbol{1}).
  \label{eq_sto_feedback_err}
\end{equation}
From Eq. \eqref{eq_sto_feedback_err}, we can see that the computational complexity of computing the gradient in our ST operation is $O(\mathcal{T})$. Moreover, for the HSPA discussed in this paper, $\boldsymbol{c}^T\boldsymbol{r}=\sum_{j\in\text{T}(\boldsymbol{s})}\boldsymbol{r}_j$, where $\text{T}(\boldsymbol{s})$ can be obtained during the inference, which makes the sublinear time complexity derived from the special structure in \eqref{eq_sto_feedback_err} efficient.
 
\section{Experiments}\label{sec:experiments}
\subsection{Datasets and Metrics}
Following previous researches \cite{zhang2018image,dai2019second,mei2021image}, we use 800 images from DIV2K \cite{timofte2017ntire} to train our deep SISR model. In a mini-batch, there are 16 images with patch size $48\times 48$ randomly cropped from the training datasets. All these training patches are augmented by random rotation of 90, 180, and 270 degrees and horizontal flipping. Then, the SR performance of our model is evaluated on five standard SISR benchmarks: Set5\cite{bevilacqua2012low}, Set14\cite{zeyde2010single}, B100\cite{martin2001database}, Urban100\cite{huang2015single}, and Manga109\cite{matsui2017sketch}. All results are compared by SSIM\cite{wang2004image} and PSNR metrics on the Y channel in YCbCr space.
\subsection{Training Details}
Our deep high-similarity-pass attention network (HSPAN) is constructed by 10 high-similarity-pass attention-based modules (HSPAMs) in a residual backbone. The number of residual blocks in locality exploration block (LEB) is set to 4 empirically. All the convolutional kernel sizes are set to $3 \times 3$, except those in our high-similarity-pass attention (HSPA), which have $1 \times 1$ kernel sizes. Intermediate features have 192 channels, and the last convolution layer in our HSPAN has 3 filters to transfer deep features into a 3-channel RGB image.

During training, the ADAM algorithm \cite{kingad2015methodforstochasticoptimization} with $\beta_1= 0.9$, $\beta_2= 0.999$, and  $\epsilon=10^{-8}$ is used for optimizing the mean absolute error (MAE) loss. The initial learning rate is set to $10^{-4}$ in $\times 2$ model and reduced to half every 200 epochs until the training stops at 1000 epochs. The $\times 3$ and $\times 4$ models are initialized by the pre-trained $\times 2$ model, and the learning rate $10^{-4}$ is reduced to half every 50 epochs until the fine-tunning stops at 200 epochs. All our models are implemented by PyTorch and trained on Nvidia 3090 GPUs.

\subsection{Ablation Studies}
We conducted an in-depth analysis of the proposed HSPA and LEB in ablation studies and trained our HSPAN on DIV2K\cite{timofte2017ntire} for classical SISR with scale factor $\times 2$. The best PSNR (dB) values on Set5\cite{bevilacqua2012low} in $5\times10^4$ iterations are obtained for comparisons.

\subsubsection{Impact of HSPA and LEB}
\ \par
The SR performance of our HSPA and LEB in the high-similarity-pass attention-based module (HSPAM) are shown in \cref{tab_ablation_core}, from which we can see that our HSPA is much better than the NLA\cite{wang2018non}. Specifically, by comparing the results in the first and second columns, our HSPA can bring 0.84dB improvement. Furthermore, when exploring locality inductive bias with the LEB, the NLA brings 0.03dB improvement, while our HSPA brings 0.11dB improvement, which is about 3.7 times that of the NLA.

\begin{table}[!htbp]
\centering
\caption{Ablation studeis on our HSPA and LEB. Best and second best results are \textbf{highlighted} and \underline{underlined}.}
\label{tab_ablation_core}
\begin{tabular}{|c|c|c|c|c|c|}
\hline
\multirow{2}{*}{LEB} & \multirow{2}{*}{\XSolidBrush} & \multirow{2}{*}{\XSolidBrush} & \multirow{2}{*}{\Checkmark} & \multirow{2}{*}{\Checkmark} & \multirow{2}{*}{\Checkmark} \\
     &       &       &       &             &                \\ \hline
\multirow{2}{*}{NLA\cite{wang2018non}} & \multirow{2}{*}{\Checkmark} & \multirow{2}{*}{\XSolidBrush} & \multirow{2}{*}{\XSolidBrush} & \multirow{2}{*}{\Checkmark} & \multirow{2}{*}{\XSolidBrush} \\
     &       &       &       &             &                \\ \hline
\multirow{2}{*}{HSPA} & \multirow{2}{*}{\XSolidBrush} & \multirow{2}{*}{\Checkmark} & \multirow{2}{*}{\XSolidBrush} & \multirow{2}{*}{\XSolidBrush} & \multirow{2}{*}{\Checkmark} \\
     &       &       &       &             &                \\ \hline
\multirow{2}{*}{PSNR} & \multirow{2}{*}{36.69} & \multirow{2}{*}{37.53} & \multirow{2}{*}{37.90 (Backbone)} & \multirow{2}{*}{\underline{37.93}} & \multirow{2}{*}{\textbf{38.01}} \\
     &       &       &       &             &                \\ \hline
\end{tabular}
\end{table}

\begin{figure*}[htbp]
\centering
\hspace{5mm}
\begin{minipage}[]{0.18\textwidth}
\begin{subfigure}[]{1.0\textwidth}
\centering
\includegraphics[height=1.415\textwidth,width=1.30\textwidth]{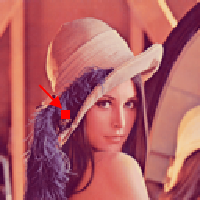}
\end{subfigure}
\end{minipage}
\begin{minipage}[]{0.78\textwidth}
\rotatebox{90}{\parbox[c]{6mm}{NLA}}
\centering
	\begin{subfigure}[]{0.16\textwidth}
		\centering
		%\vspace{-15mm}
		\includegraphics[width=\textwidth, height=\textwidth]{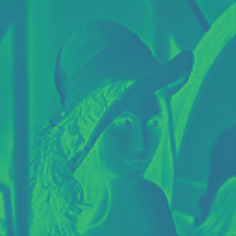}
	\end{subfigure}
%	\begin{subfigure}[]{0.09\textwidth}
%		\centering
%		%\vspace{-15mm}
%		\includegraphics[width=\textwidth, height=\textwidth]{images/fea_cmp/NLAN_C192R2_x4_tmp_1.png}
%	\end{subfigure}
	\begin{subfigure}[]{0.16\textwidth}
		\centering
		%\vspace{-15mm}
		\includegraphics[width=\textwidth, height=\textwidth]{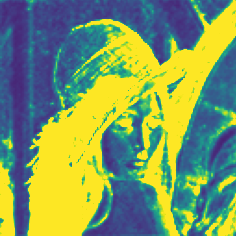}
	\end{subfigure}
%	\begin{subfigure}[]{0.09\textwidth}
%		\centering
%		%\vspace{-15mm}
%		\includegraphics[width=\textwidth, height=\textwidth]{images/fea_cmp/NLAN_C192R2_x4_tmp_3.png}
%	\end{subfigure}
	\begin{subfigure}[]{0.16\textwidth}
		\centering
		%\vspace{-15mm}
		\includegraphics[width=\textwidth, height=\textwidth]{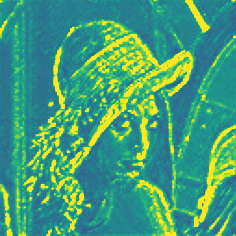}
	\end{subfigure}
%	\begin{subfigure}[]{0.09\textwidth}
%		\centering
%		%\vspace{-15mm}
%		\includegraphics[width=\textwidth, height=\textwidth]{images/fea_cmp/NLAN_C192R2_x4_tmp_5.png}
%	\end{subfigure}
	\begin{subfigure}[]{0.16\textwidth}
		\centering
		%\vspace{-15mm}
		\includegraphics[width=\textwidth, height=\textwidth]{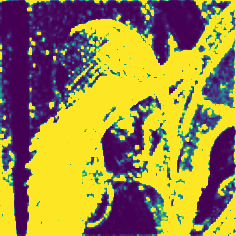}
	\end{subfigure}
%	\begin{subfigure}[]{0.09\textwidth}
%		\centering
%		%\vspace{-15mm}
%		\includegraphics[width=\textwidth, height=\textwidth]{images/fea_cmp/NLAN_C192R2_x4_tmp_7.png}
%	\end{subfigure}
	\begin{subfigure}[]{0.16\textwidth}
		\centering
		%\vspace{-15mm}
		\includegraphics[width=\textwidth, height=\textwidth]{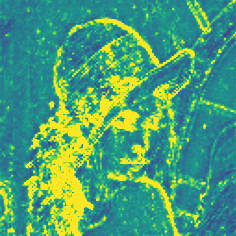}
	\end{subfigure}
%	\begin{subfigure}[]{0.09\textwidth}
%		\centering
%		%\vspace{-15mm}
%		\includegraphics[width=\textwidth, height=\textwidth]{images/fea_cmp/NLAN_C192R2_x4_tmp_9.png}
%	\end{subfigure}
%\end{minipage}

\vspace{2mm}
%
%\begin{minipage}[]{\textwidth}
%\centering

\rotatebox{90}{\parbox[c]{6mm}{HSPA}}
	\begin{subfigure}[]{0.16\textwidth}
		\centering
		\vspace{-1mm}
		\includegraphics[width=\textwidth, height=\textwidth]{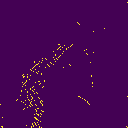}
	\end{subfigure}
%	\begin{subfigure}[]{0.09\textwidth}
%		\centering
%		\vspace{-1mm}
%		\includegraphics[width=\textwidth, height=\textwidth]{images/fea_cmp/SNLAN_C192R2K128_x4_tmp_1.png}
%	\end{subfigure}
	\begin{subfigure}[]{0.16\textwidth}
		\centering
		\vspace{-1mm}
		\includegraphics[width=\textwidth, height=\textwidth]{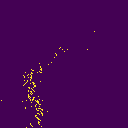}
	\end{subfigure}
%	\begin{subfigure}[]{0.09\textwidth}
%		\centering
%		\vspace{-1mm}
%		\includegraphics[width=\textwidth, height=\textwidth]{images/fea_cmp/SNLAN_C192R2K128_x4_tmp_3.png}
%	\end{subfigure}
	\begin{subfigure}[]{0.16\textwidth}
		\centering
		\vspace{-1mm}
		\includegraphics[width=\textwidth, height=\textwidth]{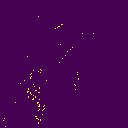}
	\end{subfigure}
%	\begin{subfigure}[]{0.09\textwidth}
%		\centering
%		\vspace{-1mm}
%		\includegraphics[width=\textwidth, height=\textwidth]{images/fea_cmp/SNLAN_C192R2K128_x4_tmp_5.png}
%	\end{subfigure}
	\begin{subfigure}[]{0.16\textwidth}
		\centering
		\vspace{-1mm}
		\includegraphics[width=\textwidth, height=\textwidth]{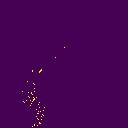}
	\end{subfigure}
%	\begin{subfigure}[]{0.09\textwidth}
%		\centering
%		\vspace{-1mm}
%		\includegraphics[width=\textwidth, height=\textwidth]{images/fea_cmp/SNLAN_C192R2K128_x4_tmp_7.png}
%	\end{subfigure}
	\begin{subfigure}[]{0.16\textwidth}
		\centering
		\vspace{-1mm}
		\includegraphics[width=\textwidth, height=\textwidth]{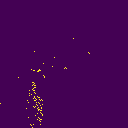}
	\end{subfigure}
%	\begin{subfigure}[]{0.09\textwidth}
%		\centering
%		\vspace{-1mm}
%		\includegraphics[width=\textwidth, height=\textwidth]{images/fea_cmp/SNLAN_C192R2K128_x4_tmp_9.png}
%	\end{subfigure}
\end{minipage}	
\centering
      \caption{Comparisons between our HSPA and the NLA on attention maps for x4 SR. These attention maps from left to right correspond to the 1st, 3rd, 5th, 7th, and 9th attention operations in our deep SISR backbone. Please zoom in for best view.}
	\label{fig_att_map_cmp}
\end{figure*}

\begin{figure}[!htbp]
\centering
\begin{minipage}[]{0.4\textwidth}
\centering

	\begin{subfigure}[]{0.3\textwidth}
		\centering
		\includegraphics[width=\textwidth]{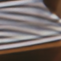}
		\vspace{-2.8mm}
	\end{subfigure}
	\begin{subfigure}[]{0.3\textwidth}
		\centering
		\includegraphics[width=\textwidth]{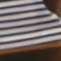}
		\vspace{-2.8mm}
	\end{subfigure}
	\begin{subfigure}[]{0.3\textwidth}
		\centering
		\includegraphics[width=\textwidth]{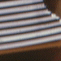}
		\vspace{-2.8mm}
	\end{subfigure}

	\begin{subfigure}[]{0.3\textwidth}
		\centering
		\includegraphics[width=\textwidth]{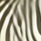}
		\vspace{-2.8mm}
	\end{subfigure}
	\begin{subfigure}[]{0.3\textwidth}
		\centering
		\includegraphics[width=\textwidth]{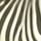}
		\vspace{-2.8mm}
	\end{subfigure}
	\begin{subfigure}[]{0.3\textwidth}
		\centering
		\includegraphics[width=\textwidth]{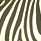}
		\vspace{-2.8mm}
	\end{subfigure}

	\begin{subfigure}[]{0.3\textwidth}
		\centering
		\includegraphics[width=\textwidth]{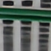}
		\subcaption*{NLA}
	\end{subfigure}
	\begin{subfigure}[]{0.3\textwidth}
		\centering
		\includegraphics[width=\textwidth]{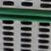}
		\subcaption*{HSPA (Ours)}
	\end{subfigure}
	\begin{subfigure}[]{0.3\textwidth}
		\centering
		\includegraphics[width=\textwidth]{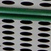}
		\subcaption*{HR}
	\end{subfigure}

\end{minipage}
\centering
      \caption{Visual comparisons between the NLA\cite{wang2018non} with our HSPA for $\times 4$ SR. The texture regions from top to bottom belong to the 'barbara', '253027' and 'img\_004' images from Set14\cite{zeyde2010single}, B100\cite{martin2001database} and  Urban100\cite{huang2015single} datasets, respectively.}
	\label{fig:nla_snla_comp}
\end{figure}

In \cref{tab_ablation_core}, we can find that the SR performance will degrade severely when the LEB is removed. It means that the locality inductive bias is important for SR reconstruction. Thus, to benifit from both locality inductive bias and non-local information, we combine the LEB and HSPA to construct our HSPAM.
%\begin{table}[!htbp]
%\centering
%\caption{Ablation studeis on our SNLA and LEB. Best and second best results are \textbf{highlighted} and \underline{underlined}.}
%\label{tab_ablation_core}
%\begin{tabular}{|c|c|c|c|c|c|}
%\hline
%LEB           & \XSolidBrush     & \XSolidBrush     & \Checkmark  & \Checkmark & \Checkmark  \\ \hline
%NLA\cite{wang2018non}    & \Checkmark  & \XSolidBrush     & \XSolidBrush     & \Checkmark & \XSolidBrush     \\ \hline
%SNLA & \XSolidBrush     & \Checkmark  & \XSolidBrush     & \XSolidBrush    & \Checkmark  \\ \hline
%PSNR           & 36.69 & 37.53 & 37.90 & \underline{37.93} & \textbf{38.01} \\ \hline
%\end{tabular}
%\end{table}

Visual comparisons between the NLA and our HSPA are shown in \cref{fig:nla_snla_comp}, from which we can compare the zoomed in results on Set14\cite{zeyde2010single}, B100\cite{martin2001database} and Urban100\cite{huang2015single} datasets. In \cref{fig:nla_snla_comp}, we can observe that our HSPA can generate superior textures than the NLA. For example, in image 'img\_004' (bottom) from Urban100, our HSPA can recover the severely damaged black circles missed by the NLA. Furthermore, we show the attention map of the NLA and our HSPA at some attention layers in \cref{fig_att_map_cmp}, from which we can observe that our HSPA can remove irrelevant features, while the NLA must has full support for every non-local feature. These attention maps indicate that our HSPA indeed works as expected: providing more relevant and informative non-local textures for SISR reconstruction with the sparse representation.

%\subsubsection{Impact of Channel Number}
%\ \par
%As can be seen from \textcolor{red}{Fig. xxx}, the SR performance keeps increasing for large channel numbers. Therefore, to pursue the SR reconstruction quality, we set the number of channels in our final DLSN model to 256.
\subsubsection{Impact of Top-k}\label{section_topk}
\begin{figure}[t]
\centering
\hspace{-7mm}
  \includegraphics[width=\linewidth]{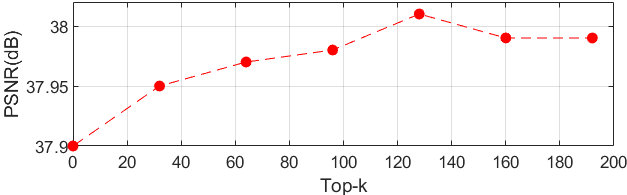}
   \caption{The PSNR results of different $k$ setting.}
   \label{top_k_psnr}
\end{figure}

As discussed in the \cref{snla}, $k$ determines the search space of the query feature, which consists of the top-k similar non-local features. The value of $k$ can be set flexibly in testing phase to find the trade-offs between reducing computational complexity and getting accurate reconstruction results. $k=0$ corresponds to the case where HSPA is not used. The SR performance of different $k$ setting are shown in \cref{top_k_psnr}, from which we can see that the SR performance of our HSPAN peaks at $k=128$. Considering the computational complexity and reconstruction performance, we set $k$ to be 128 in our HSPAN.

\subsubsection{Generic validation}
\ \par
We validated the generality of the proposed method through two parts: (1) we directly integrated the proposed HSPA into SISR models that did not use non-local attention (NLA); (2) we replaced the softmax transformation in NLA-based deep SISR models with our soft thresholding (ST) operation. 

\begin{figure}[!htbp]
  \centering
  \includegraphics[width=0.97\linewidth]{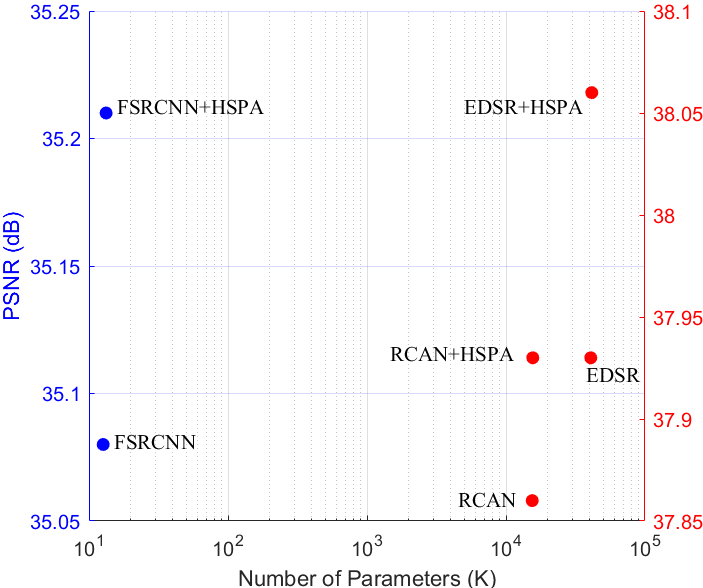}
   \caption{Parameters vs. performance. Our HSPA can improve the SR performance of representative deep SISR models vary in complexity from the simple FSRCNN\cite{dong2016accelerating} to the very complex EDSR\cite{lim2017enhanced} and RCAN\cite{zhang2018image}.}
   \label{fig_versatility}
\end{figure}

\noindent\textbf{Generic of the HSPA.} To demonstrate the generic of our HSPA, we inserted the HSPA into some representative deep SISR models with different computational complexity, such as FSRCNN\cite{dong2016accelerating}, EDSR\cite{lim2017enhanced}, and RCAN\cite{zhang2018image}. From \cref{fig_versatility}, we observe that our HSPA can significantly improve the SR performance of these deep SISR models with negligible parameters. For example, our HSPA brings 0.13dB improvement for FSRCNN\cite{dong2016accelerating} (12644 parameters) with only 630 parameters.

\begin{figure}[!htbp]
  \centering
  \includegraphics[width=\linewidth]{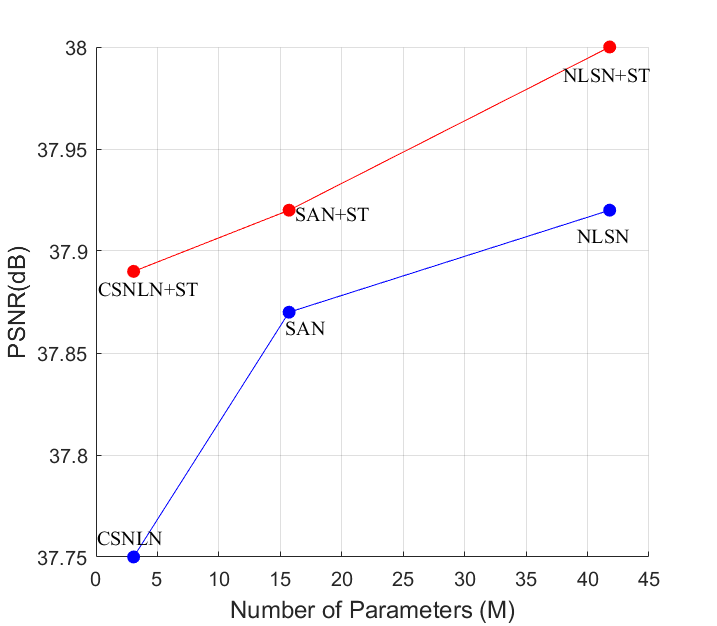}
   \caption{Parameters vs. performance. The SR performance of typical NLA-based deep SISR models can be improved by our soft thresholding (ST) operation without introducing additional parameters.}
   \label{fig_versatility1}
\end{figure}

\noindent\textbf{Generic of the ST operation.} In NLA-based deep SISR models, we replaced the softmax transformation with our ST operation in considerable CSNLN\cite{mei2020image}, NLSN\cite{mei2021image}, and SAN\cite{dai2019second}, respectively. As shown in \cref{fig_versatility1}, we can see that our ST operation can improve the SR performance of NLA-based deep SISR models without additional parameters. Specifically, our ST operation brought performance improvements of 0.14dB, 0.08dB, and 0.05dB for CSNLN, NLSN and SAN, respectively. Consistent with subjective perception, we found that the more NLA modules used in NLA-based deep SISR models, the more significant improvement our ST operation could bring after replacing the softmax transformation. For example, CSNLN used 24 NLA modules, and our ST operation brought the most significant improvement to this model, while SAN only used two NLA modules, resulting in relatively limited performance improvement. These results indicate that the proposed method can be integrated into existing deep SISR models as an effective general building unit.


\subsection{Model Efficiency}
In this subsection, we compare the efficiency of our HSPAN with other state-of-the-art models in terms of model parameters and inference time.

\begin{figure}[!htbp]
  \centering
  \includegraphics[width=0.9\linewidth]{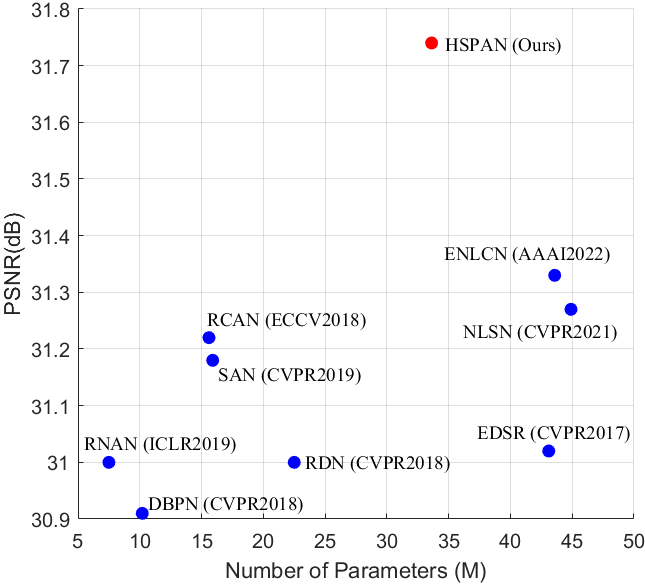}
   \caption{Model parameters and SR performance comparsions on Manga109 ($\times 4$).}
   \label{fig_parameters_psnr}
\end{figure}

\noindent\textbf{Model Parameters.} The parameters and SR performance of state-of-the-art deep SISR models including EDSR\cite{lim2017enhanced}, RDN\cite{zhang2018residual}, RCAN\cite{zhang2018image}, DBPN\cite{haris2018deep}, RNAN\cite{zhang2019residual}, SAN\cite{dai2019second}, NLSN\cite{mei2021image}, ENLCN\cite{xia2022efficient} are shown in \cref{fig_parameters_psnr}, from which we can see that the SR performance of our HSPAN on Manga109 ($\times 4$) is significantly better than other state-of-the-art models. Specifically, our HSPAN (about 33.6M parameters) brings 0.47dB improvement in SR performance with much lower parameters than the recent state-of-the-art NLSN\cite{mei2021image} (about 44.9M parameters). In contrast, compared with RCAN, NLSN brings 0.05dB improvement with 29.3M additional parameters. This means that the proposed HSPAN can indeed significantly improve the SR performance and the improvement is not simply the result of having more parameters in the network.

\begin{figure}[!htbp]
	\centering
	\includegraphics[height=0.39\textwidth]{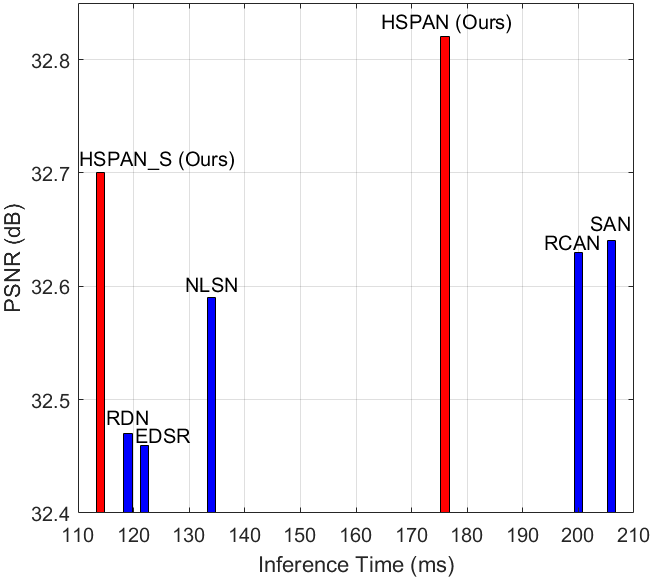}
   \caption{Inference time and SR performance comparisions on Set5 ($\times 4$).}
	\label{fig_time_psnr}
\end{figure}
\begin{table*}[!htbp]
\begin{center}
\caption{Quantitative results on SISR benchmark datasets. Best and second best results are \textbf{highlighted} and \underline{underlined}.}
\label{tab:x2_x3_x4_psnr_ssim}
\resizebox{0.78\textwidth}{!}{
\begin{tabular}{|c|c|cc|cc|cc|cc|cc|}
\hline
\multirow{2}{*}{Method} &
  \multirow{2}{*}{Scale} &
  \multicolumn{2}{c|}{Set5\cite{bevilacqua2012low}} &
  \multicolumn{2}{c|}{Set14\cite{zeyde2010single}} &
  \multicolumn{2}{c|}{B100\cite{martin2001database}} &
  \multicolumn{2}{c|}{Urban100\cite{huang2015single}} &
  \multicolumn{2}{c|}{Manga109\cite{matsui2017sketch}} \\ \cline{3-12} 
 &
   &
  \multicolumn{1}{c|}{PSNR} &
  SSIM &
  \multicolumn{1}{c|}{PSNR} &
  SSIM &
  \multicolumn{1}{c|}{PSNR} &
  SSIM &
  \multicolumn{1}{c|}{PSNR} &
  SSIM &
  \multicolumn{1}{c|}{PSNR} &
  SSIM \\ \hline

\begin{tabular}[c]{@{}c@{}}Bicubic\\ FSRCNN\cite{dong2016accelerating}\\ VDSR\cite{kim2016accurate}\\ LapSRN\cite{lai2018fast}\\ MemNet\cite{tai2017memnet}\\ SRMDNF\cite{zhang2018learning}\\ DBPN\cite{haris2018deep}\\ EDSR\cite{lim2017enhanced}\\ RDN\cite{zhang2018residual}\\ RCAN\cite{zhang2018image}\\ SAN\cite{dai2019second}\\ OISR\cite{he2019ode}\\ IGNN\cite{zhou2020cross}\\ CSNLN\cite{mei2020image}\\ HAN\cite{niu2020single}\\  NLSN\cite{mei2021image}\\ DRLN\cite{anwar2022densely}\\ SwinIR\cite{liang2021swinir}\\ ENLCN\cite{xia2022efficient}\end{tabular} &
  \begin{tabular}[c]{@{}c@{}}$\times 2$\\ $\times 2$\\ $\times 2$\\ $\times 2$\\ $\times 2$\\ $\times 2$\\ $\times 2$\\ $\times 2$\\ $\times 2$\\ $\times 2$\\ $\times 2$\\ $\times 2$\\ $\times 2$\\ $\times 2$\\ $\times 2$\\ $\times 2$\\ $\times 2$\\ $\times 2$\\ $\times 2$\end{tabular} &
  \multicolumn{1}{c|}{\begin{tabular}[c]{@{}c@{}}33.66\\37.05\\37.53\\37.52\\ 37.78\\ 37.79\\ 38.09\\ 38.11\\38.24\\ 38.27\\ 38.31\\ 38.21\\ 38.24\\ 38.28\\ 38.27\\ 38.34\\ 38.27\\ 38.35\\ 38.37\end{tabular}} &
  \begin{tabular}[c]{@{}c@{}}0.9299\\0.9560\\0.9590\\0.9591\\ 0.9597\\ 0.9601\\ 0.9600\\ 0.9602\\ 0.9614\\ 0.9614\\ 0.9620\\ 0.9612\\ 0.9613\\ 0.9616\\ 0.9614\\ 0.9618\\ 0.9616\\ 0.9620\\ 0.9618\end{tabular} &
  \multicolumn{1}{c|}{\begin{tabular}[c]{@{}c@{}}30.24\\32.66\\33.05\\33.08\\ 33.28\\ 33.32\\ 33.85\\ 33.92\\ 34.01\\ 34.12\\ 34.07\\ 33.94\\ 34.12\\ 34.07\\ 34.16\\ 34.08\\ 34.28\\ 34.14\\ 34.17\end{tabular}} &
  \begin{tabular}[c]{@{}c@{}}0.8688\\0.9090\\0.9130\\0.9130\\ 0.9142\\ 0.9159\\ 0.9190\\ 0.9195\\ 0.9212\\ 0.9216\\ 0.9213\\ 0.9206\\ 0.9217\\ 0.9223\\ 0.9217\\ 0.9231\\ 0.9231\\ 0.9227\\ 0.9229\end{tabular} &
  \multicolumn{1}{c|}{\begin{tabular}[c]{@{}c@{}}29.56\\31.53\\31.90\\31.08\\ 32.08\\ 32.05\\ 32.27\\ 32.32\\ 32.34\\ 32.41\\ 32.42\\ 32.36\\ 32.41\\ 32.40\\ 32.41\\ 32.43\\ 32.44\\ 32.44\\ \underline{32.49}\end{tabular}} &
  \begin{tabular}[c]{@{}c@{}}0.8431\\0.8920\\0.8960\\0.8950\\ 0.8978\\ 0.8985\\ 0.9000\\ 0.9013\\ 0.9017\\ 0.9027\\ 0.9028\\ 0.9019\\ 0.9025\\ 0.9024\\ 0.9027\\ 0.9027\\ 0.9028\\ 0.9030\\ 0.9032\end{tabular} &
  \multicolumn{1}{c|}{\begin{tabular}[c]{@{}c@{}}26.88\\29.88\\30.77\\30.41\\ 31.31\\ 31.33\\ 32.55\\ 32.93\\ 32.89\\ 33.34\\ 33.10\\ 33.03\\ 33.23\\ 33.25\\ 33.35\\ 33.42\\ 33.37\\ 33.40\\ 33.56\end{tabular}} &
  \begin{tabular}[c]{@{}c@{}}0.8403\\0.9020\\0.9140\\0.9101\\ 0.9195\\ 0.9204\\ 0.9324\\ 0.9351\\ 0.9353\\ 0.9384\\ 0.9370\\ 0.9365\\ 0.9383\\ 0.9386\\ 0.9385\\ 0.9394\\ 0.9390\\ 0.9393\\ 0.9398\end{tabular} &
  \multicolumn{1}{c|}{\begin{tabular}[c]{@{}c@{}}30.80\\36.67\\37.22\\37.27\\ 37.72\\ 38.07\\ 38.89\\ 39.10\\ 39.18\\ 39.44\\ 39.32\\ --\\ 39.35\\ 39.37\\ 39.46\\ 39.59\\ 39.58\\ 39.60\\ 39.64\end{tabular}} &
  \begin{tabular}[c]{@{}c@{}}0.9339\\0.9710\\0.9750\\0.9740\\ 0.9740\\ 0.9761\\ 0.9775\\ 0.9773\\ 0.9780\\ 0.9786\\ 0.9792\\ --\\ 0.9786\\ 0.9785\\ 0.9785\\ 0.9789\\ 0.9786\\ \underline{0.9792}\\ 0.9791
\end{tabular} \\ \hline
\begin{tabular}[c]{@{}c@{}}HSPAN (Ours)\\ HSPAN+ (Ours)\end{tabular} &
  \begin{tabular}[c]{@{}c@{}}$\times 2$\\ $\times 2$\end{tabular} &
  \multicolumn{1}{c|}{\begin{tabular}[c]{@{}c@{}}\underline{38.43}\\ \textbf{38.51}\end{tabular}} &
  \begin{tabular}[c]{@{}c@{}}\underline{0.9622}\\ \textbf{0.9625}\end{tabular} &
  \multicolumn{1}{c|}{\begin{tabular}[c]{@{}c@{}}\underline{34.30}\\ \textbf{34.47}\end{tabular}} &
  \begin{tabular}[c]{@{}c@{}}\underline{0.9246}\\ \textbf{0.9250}\end{tabular} &
  \multicolumn{1}{c|}{\begin{tabular}[c]{@{}c@{}}32.48\\ \textbf{32.53}\end{tabular}} &
  \begin{tabular}[c]{@{}c@{}}\underline{0.9037}\\ \textbf{0.9043}\end{tabular} &
  \multicolumn{1}{c|}{\begin{tabular}[c]{@{}c@{}}\underline{33.87}\\ \textbf{34.09}\end{tabular}} &
  \begin{tabular}[c]{@{}c@{}}\underline{0.9427}\\ \textbf{0.9441}\end{tabular} &
  \multicolumn{1}{c|}{\begin{tabular}[c]{@{}c@{}}\underline{39.67}\\ \textbf{39.87}\end{tabular}} &
  \begin{tabular}[c]{@{}c@{}}0.9787\\ \textbf{0.9793}\end{tabular} \\ \hline
\begin{tabular}[c]{@{}c@{}}Bicubic\\ FSRCNN\cite{dong2016accelerating}\\ VDSR\cite{kim2016accurate}\\ LapSRN\cite{lai2018fast}\\ MemNet\cite{tai2017memnet}\\ SRMDNF\cite{zhang2018learning}\\ EDSR\cite{lim2017enhanced}\\ RDN\cite{zhang2018residual}\\ RCAN\cite{zhang2018image}\\ SAN\cite{dai2019second}\\ OISR\cite{he2019ode}\\ IGNN\cite{zhou2020cross}\\ CSNLN\cite{mei2020image}\\ HAN\cite{niu2020single}\\ NLSN\cite{mei2021image}\\ DRLN\cite{anwar2022densely}\\ SwinIR\cite{liang2021swinir}\end{tabular} &
  \begin{tabular}[c]{@{}c@{}}$\times 3$\\ $\times 3$\\ $\times 3$\\ $\times 3$\\ $\times 3$\\ $\times 3$\\ $\times 3$\\ $\times 3$\\ $\times 3$\\ $\times 3$\\ $\times 3$\\ $\times 3$\\ $\times 3$\\ $\times 3$\\ $\times 3$\\ $\times 3$\\ $\times 3$\end{tabular} &
  \multicolumn{1}{c|}{\begin{tabular}[c]{@{}c@{}}30.39 \\33.18\\33.67\\33.82\\ 34.09\\ 34.12\\ 34.65\\ 34.71\\ 34.74\\ 34.75\\ 34.72\\ 34.72\\ 34.74\\ 34.75\\ 34.85\\ 34.78\\ 34.89\end{tabular}} &
  \begin{tabular}[c]{@{}c@{}}0.8682\\0.9140\\0.9210\\0.9227\\ 0.9248\\ 0.9254\\ 0.9280\\ 0.9296\\ 0.9299\\ 0.9300\\ 0.9297\\ 0.9298\\ 0.9300\\ 0.9299\\ 0.9306\\ 0.9303\\ \underline{0.9312}\end{tabular} &
  \multicolumn{1}{c|}{\begin{tabular}[c]{@{}c@{}}27.55\\29.37\\29.78\\29.87\\ 30.00\\ 30.04\\ 30.52\\ 30.57\\ 30.65\\ 30.59\\ 30.57\\ 30.66\\ 30.66\\ 30.67\\ 30.70\\ 30.73\\ \underline{30.77}\end{tabular}} &
  \begin{tabular}[c]{@{}c@{}}0.7742\\0.8240\\0.8320\\0.8320\\ 0.8350\\ 0.8382\\ 0.8462\\ 0.8468\\ 0.8482\\ 0.8476\\ 0.8470\\ 0.8484\\ 0.8482\\  0.8483\\ 0.8485\\ 0.8488\\ 0.8503\end{tabular} &
  \multicolumn{1}{c|}{\begin{tabular}[c]{@{}c@{}}27.21\\28.53\\28.83\\28.82\\ 28.96\\ 28.97\\ 29.25\\ 29.26\\ 29.32\\ 29.33\\ 29.29\\ 29.31\\ 29.33\\ 29.32\\ 29.34\\ 29.36\\ 29.37\end{tabular}} &
  \begin{tabular}[c]{@{}c@{}}0.7385\\0.7910\\0.7990\\0.7980\\ 0.8001\\ 0.8025\\ 0.8093\\ 0.8093\\ 0.8111\\ 0.8112\\ 0.8103\\ 0.8105\\ 0.8105\\ 0.8110\\ 0.8117\\ 0.8117\\ 0.8124\end{tabular} &
  \multicolumn{1}{c|}{\begin{tabular}[c]{@{}c@{}}24.46\\26.43\\27.14\\27.07\\ 27.56\\ 27.57\\ 28.80\\ 28.80\\ 29.09\\ 28.93\\ 28.95\\ 29.03\\ 29.13\\ 29.10\\ 29.25\\ 29.21\\ 29.29\end{tabular}} &
  \begin{tabular}[c]{@{}c@{}}0.7349\\0.8080\\ 0.8290\\0.8280\\ 0.8376\\ 0.8398\\ 0.8653\\ 0.8653\\ 0.8702\\ 0.8671\\ 0.8680\\ 0.8696\\ 0.8712\\ 0.8705\\ 0.8726\\ 0.8722\\ 0.8744\end{tabular} &
  \multicolumn{1}{c|}{\begin{tabular}[c]{@{}c@{}}26.95\\31.10\\32.01\\32.21\\ 32.51\\ 33.00\\ 34.17\\ 34.13\\ 34.44\\ 34.30\\ --\\ 34.39\\ 34.45\\ 34.48\\ 34.57\\ 34.71\\ 34.74\end{tabular}} &
  \begin{tabular}[c]{@{}c@{}}0.8556\\0.9210\\0.9340\\0.9350\\ 0.9369\\ 0.9403\\ 0.9476\\ 0.9484\\ 0.9499\\ 0.9494\\ --\\ 0.9496\\ 0.9502\\ 0.9500\\ 0.9508\\ 0.9509\\ 0.9518\end{tabular} \\ \hline
\begin{tabular}[c]{@{}c@{}}HSPAN (Ours) \\ HSPAN+ (Ours)\end{tabular} &
  \begin{tabular}[c]{@{}c@{}}$\times 3$\\ $\times 3$\end{tabular} &
  \multicolumn{1}{c|}{\begin{tabular}[c]{@{}c@{}}\underline{34.91}\\ \textbf{35.03}\end{tabular}} &
  \begin{tabular}[c]{@{}c@{}}0.9311\\ \textbf{0.9318}\end{tabular} &
  \multicolumn{1}{c|}{\begin{tabular}[c]{@{}c@{}}\underline{30.77}\\ \textbf{30.88}\end{tabular}} &
  \begin{tabular}[c]{@{}c@{}}\underline{0.8504}\\ \textbf{0.8518}\end{tabular} &
  \multicolumn{1}{c|}{\begin{tabular}[c]{@{}c@{}}\underline{29.41}\\ \textbf{29.46}\end{tabular}} &
  \begin{tabular}[c]{@{}c@{}}\underline{0.8137}\\ \textbf{0.8146}\end{tabular} &
  \multicolumn{1}{c|}{\begin{tabular}[c]{@{}c@{}}\underline{29.67}\\ \textbf{29.90}\end{tabular}} &
  \begin{tabular}[c]{@{}c@{}}\underline{0.8800}\\ \textbf{0.8830}\end{tabular} &
  \multicolumn{1}{c|}{\begin{tabular}[c]{@{}c@{}}\underline{34.85}\\ \textbf{35.16}\end{tabular}} &
  \begin{tabular}[c]{@{}c@{}}\underline{0.9522}\\ \textbf{0.9535}\end{tabular} \\ \hline
\begin{tabular}[c]{@{}c@{}}Bicubic\\ FSRCNN\cite{dong2016accelerating}\\ VDSR\cite{kim2016accurate}\\ LapSRN\cite{lai2018fast}\\ MemNet\cite{tai2017memnet}\\ SRMDNF\cite{zhang2018learning}\\ DBPN\cite{haris2018deep}\\ EDSR\cite{lim2017enhanced}\\ RDN\cite{zhang2018residual}\\ RCAN\cite{zhang2018image}\\ SAN\cite{dai2019second}\\ OISR\cite{he2019ode}\\ IGNN\cite{zhou2020cross}\\ CSNLN\cite{mei2020image}\\ HAN\cite{niu2020single}\\ NLSN\cite{mei2021image}\\ DRLN\cite{anwar2022densely}\\ SwinIR\cite{liang2021swinir}\\ ENLCN\cite{xia2022efficient}\end{tabular} &
  \begin{tabular}[c]{@{}c@{}}$\times 4$\\ $\times 4$\\ $\times 4$\\ $\times 4$\\ $\times 4$\\ $\times 4$\\ $\times 4$\\ $\times 4$\\ $\times 4$\\ $\times 4$\\ $\times 4$\\ $\times 4$\\ $\times 4$\\ $\times 4$\\ $\times 4$\\ $\times 4$\\ $\times 4$\\ $\times 4$\\ $\times 4$\end{tabular} &
  \multicolumn{1}{c|}{\begin{tabular}[c]{@{}c@{}}28.42\\30.72\\31.35\\31.54\\ 31.74\\ 31.96\\ 32.47\\ 32.46\\ 32.47\\ 32.63\\ 32.64\\ 32.53\\ 32.57\\ 32.68\\ 32.64\\ 32.59\\ 32.63\\ 32.72\\ 32.67\end{tabular}} &
  \begin{tabular}[c]{@{}c@{}}0.8104\\0.8660\\0.8830\\0.8850\\ 0.8893\\ 0.8925\\ 0.8980\\ 0.8968\\ 0.8990\\ 0.9002\\ 0.9003\\ 0.8992\\ 0.8998\\ 0.9004\\ 0.9002\\ 0.9000\\ 0.9002\\ \underline{0.9021}\\ 0.9004\end{tabular} &
  \multicolumn{1}{c|}{\begin{tabular}[c]{@{}c@{}}26.00\\27.61\\28.02\\28.19\\ 28.26\\ 28.35\\ 28.82\\ 28.80\\ 28.81\\ 28.87\\ 28.92\\ 28.86\\ 28.85\\ 28.95\\ 28.90\\ 28.87\\ 28.94\\ 28.94\\ 28.94\end{tabular}} &
  \begin{tabular}[c]{@{}c@{}}0.7027\\0.7550\\0.7680\\0.7720\\ 0.7723\\ 0.7787\\ 0.7860\\ 0.7876\\ 0.7871\\ 0.7889\\ 0.7888\\ 0.7878\\ 0.7891\\ 0.7888\\ 0.7890\\ 0.7891\\ 0.7900\\ 0.7914\\ 0.7892\end{tabular} &
  \multicolumn{1}{c|}{\begin{tabular}[c]{@{}c@{}}25.96\\26.98\\27.29\\27.32\\ 27.40\\ 27.49\\ 27.72\\ 27.71\\ 27.72\\ 27.77\\ 27.78\\ 27.75\\ 27.77\\ 27.80\\ 27.80\\ 27.78\\ 27.83 \\ 27.83\\ 27.82 \end{tabular}} &
  \begin{tabular}[c]{@{}c@{}}0.6675\\0.7150\\0.0726\\0.7270\\ 0.7281\\ 0.7337\\ 0.7400\\ 0.7420\\ 0.7419\\ 0.7436\\ 0.7436\\ 0.7428\\ 0.7434\\ 0.7439\\ 0.7442\\ 0.7444\\ 0.7444\\ 0.7459\\ 0.7452\end{tabular} &
  \multicolumn{1}{c|}{\begin{tabular}[c]{@{}c@{}}23.14\\24.62\\25.18\\25.21\\ 25.50\\ 25.68\\ 26.38\\ 26.64\\ 26.61\\ 26.82\\ 26.79\\ 26.79\\ 26.84\\ 27.22\\ 26.85\\ 26.96\\ 26.98\\ 27.07\\ 27.12\end{tabular}} &
  \begin{tabular}[c]{@{}c@{}}0.6577\\0.7280\\0.7540\\0.7560\\ 0.7630\\ 0.7731\\ 0.7946\\ 0.8033\\ 0.8028\\ 0.8087\\ 0.8068\\ 0.8068\\ 0.8090\\ 0.8168\\ 0.8094\\ 0.8109\\ 0.8119\\ 0.8164\\ 0.8141\end{tabular} &
  \multicolumn{1}{c|}{\begin{tabular}[c]{@{}c@{}}24.89\\27.90\\ 28.83\\29.09\\ 29.42\\ 30.09\\ 30.91\\ 31.02\\ 31.00\\ 31.22\\ 31.18\\ --\\ 31.28\\ 31.43\\ 31.42\\ 31.27\\ 31.54\\ 31.67\\ 31.33\end{tabular}} &
  \begin{tabular}[c]{@{}c@{}}0.7866\\0.8610\\0.8870\\0.8900\\ 0.8942\\ 0.9024\\ 0.9137\\ 0.9148\\ 0.9151\\ 0.9173\\ 0.9169\\ --\\ 0.9182\\ 0.9201\\ 0.9177\\ 0.9184\\ 0.9196\\ 0.9226\\ 0.9188\end{tabular} \\ \hline
\begin{tabular}[c]{@{}c@{}}HSPAN (Ours)\\ HSPAN+ (Ours)\end{tabular} &
  \begin{tabular}[c]{@{}c@{}}$\times 4$\\ $\times 4$\end{tabular} &
  \multicolumn{1}{c|}{\begin{tabular}[c]{@{}c@{}}\underline{32.82}\\ \textbf{32.94} \end{tabular}} &
  \begin{tabular}[c]{@{}c@{}}0.9013\\ \textbf{0.9027} \end{tabular} &
  \multicolumn{1}{c|}{\begin{tabular}[c]{@{}c@{}}\underline{29.03}\\ \textbf{29.12}\end{tabular}} &
  \begin{tabular}[c]{@{}c@{}}\underline{0.7920}\\ \textbf{0.7937}\end{tabular} &
  \multicolumn{1}{c|}{\begin{tabular}[c]{@{}c@{}}\underline{27.86}\\ \textbf{27.93}\end{tabular}} &
  \begin{tabular}[c]{@{}c@{}}\underline{0.7476}\\ \textbf{0.7491}\end{tabular} &
  \multicolumn{1}{c|}{\begin{tabular}[c]{@{}c@{}}\underline{27.36}\\ \textbf{27.59}\end{tabular}} &
  \begin{tabular}[c]{@{}c@{}}\underline{0.8220} \\ \textbf{0.8263}\end{tabular} &
  \multicolumn{1}{c|}{\begin{tabular}[c]{@{}c@{}}\underline{31.74}\\ \textbf{32.11}\end{tabular}} &
  \begin{tabular}[c]{@{}c@{}}\underline{0.9230} \\ \textbf{0.9257}\end{tabular} \\ \hline
\end{tabular}}
\end{center}
\end{table*}
\noindent\textbf{Inference Time.} We also provide a smaller version of our HSPAN, denoted as HSPAN\_S, by integrating only one HSPA in the middle of the backbone. The inference time of our HSPAN, HSPAN\_S and recently competitive deep SISR models are shown in \cref{fig_time_psnr}, from which we see that the SR performance of our HSPAN is much better than other state-of-the-art models. Furthermore, our HSPAN\_S consumes the least inference time while achieving better SR performance than other state-of-the-art models. Specifically, by comparing NLSN\cite{mei2021image} and RCAN\cite{zhang2018image}, we find that the SR performance of RCAN is 0.04dB higher than that of NLSN at a cost of 66 milliseconds. However, the SR performance of our HSPAN\_S is not only 0.23dB higher than that of NLSN, but also reduces the inference time by 20 milliseconds. These results indicate that our HSPA is very efficient in improving SR performance. The inferences of all models are conducted in the same environment with a Nvidia 3090 GPU.
%If we ignore the computation of SB-LSH and take an image with size $256\times 256$ as input, the computational complexity of the standard NLA is $\frac{(hw)^2c}{hwlc+hwlc+l^2c}=\frac{hw}{2l+\frac{l^2}{hw}}=\frac{256\times 256}{256+0.25}\approx256$ times that of our GLA. \textcolor{red}{Figure xxx} shows the comparison in terms of GPU memory consumption between the standard NLA and our GLA under different image input sizes. From \textcolor{red}{Figure xxx}, we can see that the GPU memory consumption of our GLA is also much lower than that of the standard NLA, and as the input size increases, the advantage of our GLA becomes more and more significant.
% Compared with the computational complexity of the standard NLA, the computation of SB-LSH is almost negligible. 
\begin{figure*}[!htbp]
\centering
\vspace{-4mm}
\hspace{10mm}\begin{minipage}[]{0.18\textwidth}
\begin{subfigure}[]{1.0\textwidth}
\centering
\vspace{-1.5mm}
\includegraphics[height=1.62\textwidth,width=1.32\textwidth]{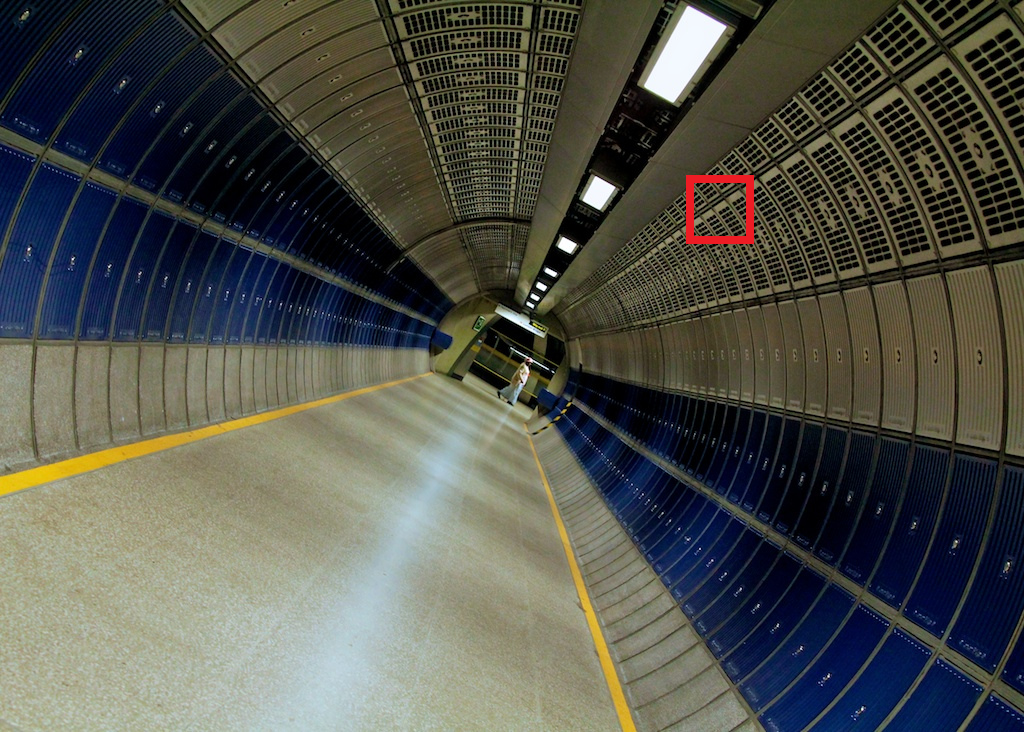}
\caption*{\ \ \ \ \ \ \ \ \ \ \ img\_078}
\end{subfigure}

\begin{subfigure}[]{1.0\textwidth}
\centering
\vspace{5.5mm}
\includegraphics[height=1.58\textwidth,width=1.32\textwidth]{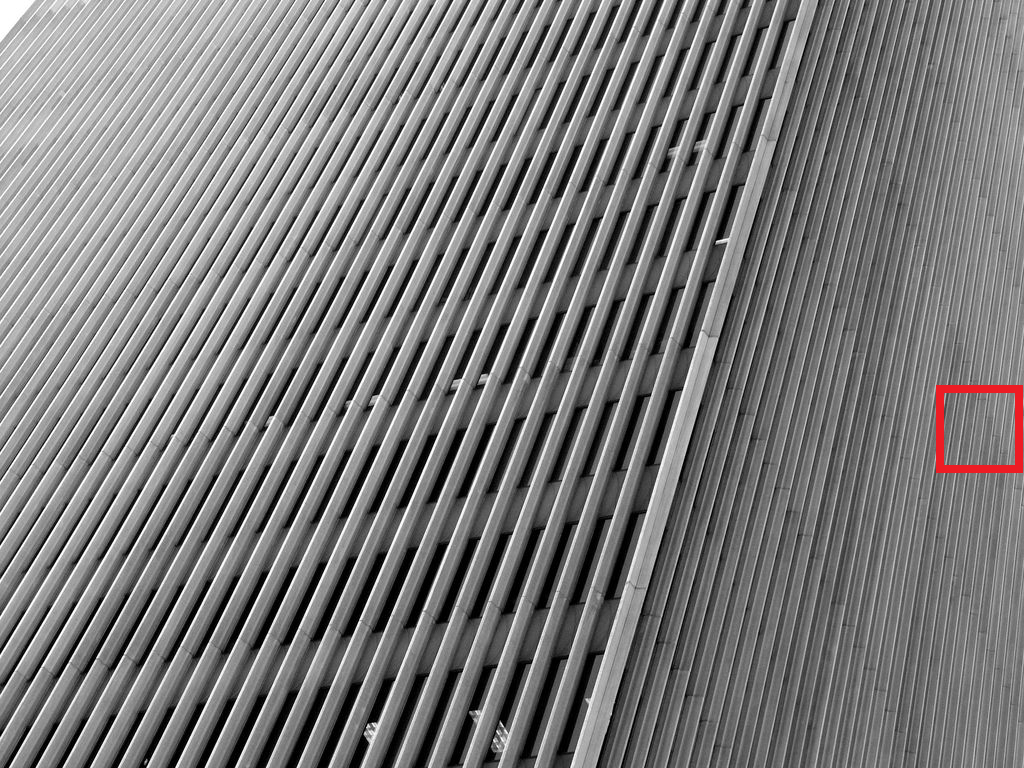}
\caption*{\ \ \ \ \ \ \ \ \ \ \ img\_045}
\end{subfigure}

\begin{subfigure}[]{1.0\textwidth}
\centering
\vspace{5.2mm}
\includegraphics[height=1.62\textwidth,width=1.35\textwidth]{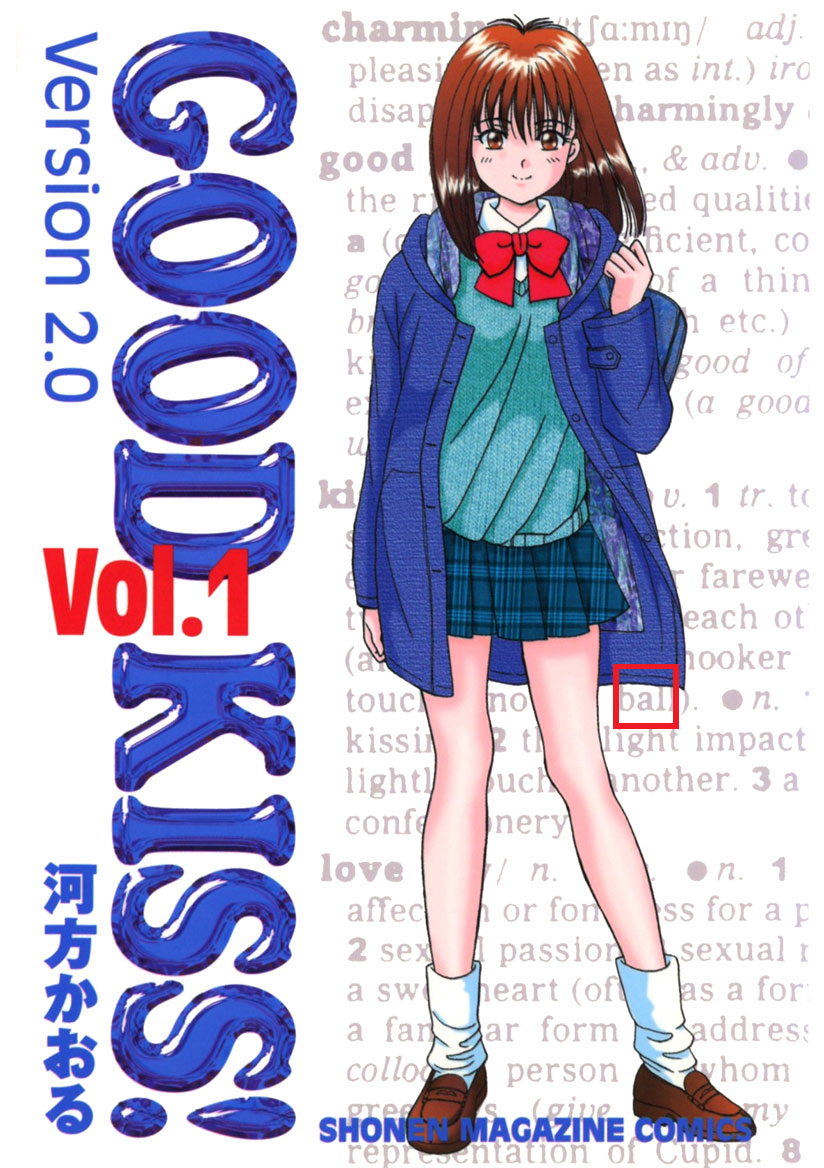}
\caption*{\parbox[c]{20mm}{GOODKISSVer2}}
\end{subfigure}

\begin{subfigure}[]{1.0\textwidth}
\centering
\vspace{5.5mm}
\includegraphics[height=1.62\textwidth,width=1.32\textwidth]{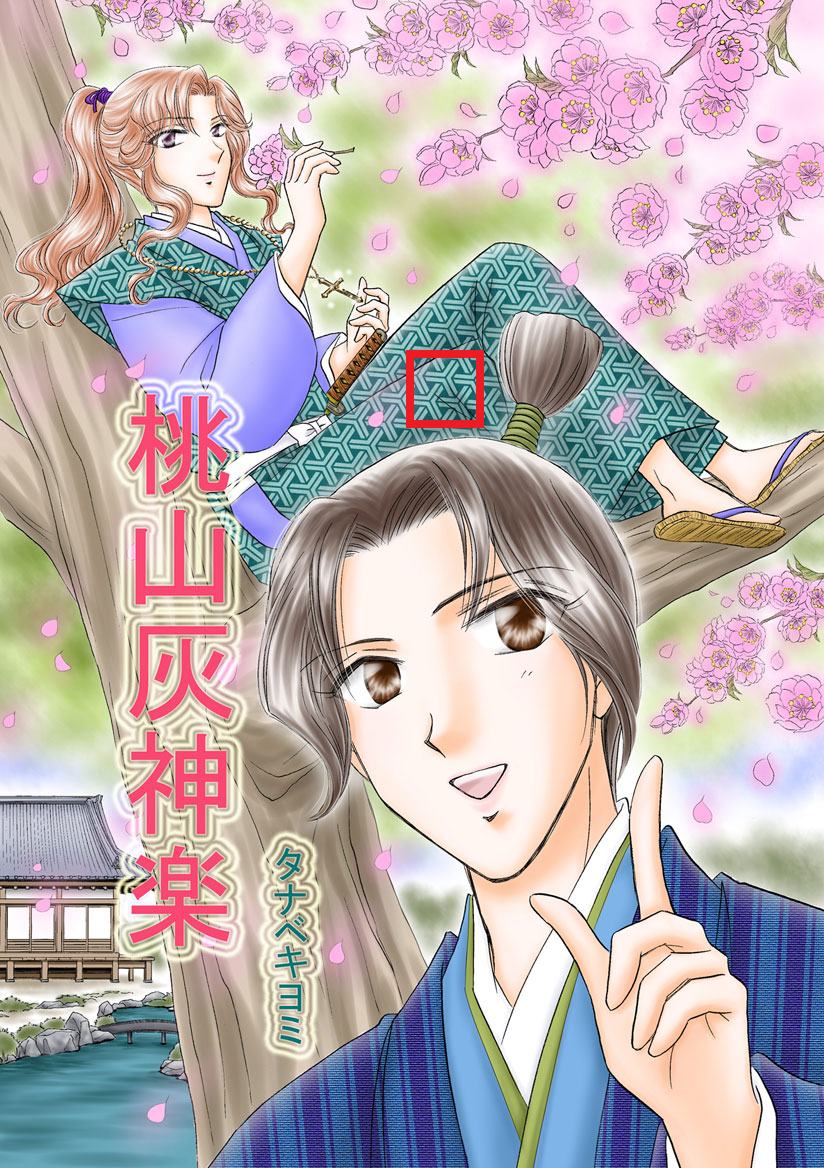}
\caption*{\parbox[c]{16mm}{MomoyamaHaikagura}}
\end{subfigure}

\end{minipage}
\hspace{-5mm}
\captionsetup[subfigure]{justification=justified,singlelinecheck=false}
\begin{minipage}[]{0.78\textwidth}
\centering
\begin{subfigure}[]{0.15\textwidth}
		\centering
		\includegraphics[width=\textwidth]{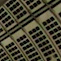}
		\caption*{\parbox[c]{6mm}{HR PSNR/SSIM}}
	\end{subfigure}
	\begin{subfigure}[]{0.15\textwidth}
		\centering
		\includegraphics[width=\textwidth]{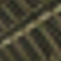}
		\caption*{\parbox[c]{8mm}{Bicubic 18.27/0.2704}}
	\end{subfigure}
	\begin{subfigure}[]{0.15\textwidth}
		\centering
		\includegraphics[width=\textwidth]{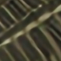}
		\caption*{\parbox[c]{15mm}{VDSR\cite{kim2016accurate} 18.28/0.3410}}
	\end{subfigure}
	\begin{subfigure}[]{0.15\textwidth}
		\centering
		\includegraphics[width=\textwidth]{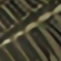}
		\caption*{\parbox[c]{18mm}{LapSRN\cite{lai2018fast} 18.32/0.3358}}
	\end{subfigure}
	\begin{subfigure}[]{0.15\textwidth}
		\centering
		\includegraphics[width=\textwidth]{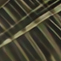}
		\caption*{\parbox[c]{15mm}{EDSR\cite{lim2017enhanced}  18.31/0.3873}}
	\end{subfigure}

\vspace{2mm}
	\begin{subfigure}[]{0.15\textwidth}
		\centering
		\includegraphics[width=\textwidth]{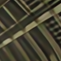}
		\caption*{\parbox[c]{16mm}{RCAN\cite{zhang2018image} 18.82/0.4252}}
	\end{subfigure}
	\begin{subfigure}[]{0.15\textwidth}
		\centering
		\includegraphics[width=\textwidth]{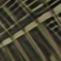}
		\caption*{\parbox[c]{15mm}{NLSN\cite{mei2021image} 19.37/0.4698}}
	\end{subfigure}
	\begin{subfigure}[]{0.15\textwidth}
		\centering
		\includegraphics[width=\textwidth]{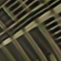}
		\caption*{\parbox[c]{16mm}{SwinIR\cite{liang2021swinir} 20.03/0.5768}}
	\end{subfigure}
	\begin{subfigure}[]{0.15\textwidth}
		\centering
		\includegraphics[width=\textwidth]{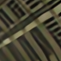}
		\caption*{\parbox[c]{17mm}{ENLCN\cite{xia2022efficient} 19.22/0.4646}}
	\end{subfigure}
	\begin{subfigure}[]{0.15\textwidth}
		\centering
		\includegraphics[width=\textwidth]{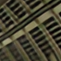}
		\caption*{\parbox[c]{10mm}{\textbf{HSPAN(Ours) 21.46/0.7382}}}
	\end{subfigure}

\vspace{4mm}
\begin{subfigure}[]{0.15\textwidth}
		\centering
		\includegraphics[width=\textwidth]{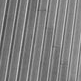}
		\caption*{\parbox[c]{6mm}{HR PSNR/SSIM}}
	\end{subfigure}
	\begin{subfigure}[]{0.15\textwidth}
		\centering
		\includegraphics[width=\textwidth]{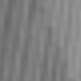}
		\caption*{\parbox[c]{8mm}{Bicubic 22.99/0.2856}}
	\end{subfigure}
	\begin{subfigure}[]{0.15\textwidth}
		\centering
		\includegraphics[width=\textwidth]{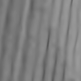}
		\caption*{\parbox[c]{15mm}{VDSR\cite{kim2016accurate} 23.04/0.3346}}
	\end{subfigure}
	\begin{subfigure}[]{0.15\textwidth}
		\centering
		\includegraphics[width=\textwidth]{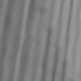}
		\caption*{\parbox[c]{18mm}{LapSRN\cite{lai2018fast} 23.20/0.3457}}
	\end{subfigure}
	\begin{subfigure}[]{0.15\textwidth}
		\centering
		\includegraphics[width=\textwidth]{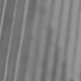}
		\caption*{\parbox[c]{15mm}{EDSR\cite{lim2017enhanced}  23.57/0.3848}}
	\end{subfigure}
	
\vspace{1mm}
	\begin{subfigure}[]{0.15\textwidth}
		\centering
		\includegraphics[width=\textwidth]{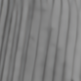}
		\caption*{\parbox[c]{16mm}{RCAN\cite{zhang2018image} 23.33/0.3827}}
	\end{subfigure}
	\begin{subfigure}[]{0.15\textwidth}
		\centering
		\includegraphics[width=\textwidth]{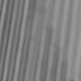}
		\caption*{\parbox[c]{15mm}{NLSN\cite{mei2021image} \textbf{23.72}/0.4011}}
	\end{subfigure}
	\begin{subfigure}[]{0.15\textwidth}
		\centering
		\includegraphics[width=\textwidth]{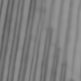}
		\caption*{\parbox[c]{16mm}{SwinIR\cite{liang2021swinir} 23.34/0.3437}}
	\end{subfigure}
	\begin{subfigure}[]{0.15\textwidth}
		\centering
		\includegraphics[width=\textwidth]{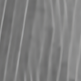}
		\caption*{\parbox[c]{17mm}{ENLCN\cite{xia2022efficient} 23.64/0.3912}}
	\end{subfigure}
	\begin{subfigure}[]{0.15\textwidth}
		\centering
		\includegraphics[width=\textwidth]{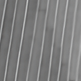}
		\caption*{\parbox[c]{10mm}{\textbf{HSPAN(Ours)} 23.49/\textbf{0.4648}}}
	\end{subfigure}

\centering
	\vspace{3mm}
	\begin{subfigure}[]{0.15\textwidth}
		\centering
		\includegraphics[width=\textwidth]{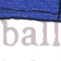}
		\caption*{\parbox[c]{8mm}{HR PSNR/SSIM}}
	\end{subfigure}
	\begin{subfigure}[]{0.15\textwidth}
		\centering
		\includegraphics[width=\textwidth]{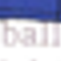}
		\caption*{\parbox[c]{8mm}{Bicubic 23.58/0.5820}}
	\end{subfigure}
	\begin{subfigure}[]{0.15\textwidth}
		\centering
		\includegraphics[width=\textwidth]{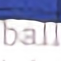}
		\caption*{\parbox[c]{15mm}{VDSR\cite{kim2016accurate} 26.62/0.7700}}
	\end{subfigure}
	\begin{subfigure}[]{0.15\textwidth}
		\centering
		\includegraphics[width=\textwidth]{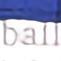}
		\caption*{\parbox[c]{18mm}{LapSRN\cite{lai2018fast} 26.61/0.7814}}
	\end{subfigure}
	\begin{subfigure}[]{0.15\textwidth}
		\centering
		\includegraphics[width=\textwidth]{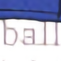}
		\caption*{\parbox[c]{15mm}{EDSR\cite{lim2017enhanced} 26.96/0.8261}}
	\end{subfigure}

\vspace{2mm}
	\begin{subfigure}[]{0.15\textwidth}
		\centering
		\includegraphics[width=\textwidth]{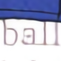}
		\caption*{\parbox[c]{15mm}{RCAN\cite{zhang2018image} 26.51/0.8190}}
	\end{subfigure}
	\begin{subfigure}[]{0.15\textwidth}
		\centering
		\includegraphics[width=\textwidth]{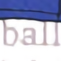}
		\caption*{\parbox[c]{15mm}{NLSN\cite{mei2021image} 30.67/0.9080}}
	\end{subfigure}
	\begin{subfigure}[]{0.15\textwidth}
		\centering
		\includegraphics[width=\textwidth]{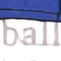}
		\caption*{\parbox[c]{16mm}{SwinIR\cite{liang2021swinir} 30.95/0.9058}}
	\end{subfigure}
	\begin{subfigure}[]{0.15\textwidth}
		\centering
		\includegraphics[width=\textwidth]{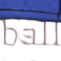}
		\caption*{\parbox[c]{17mm}{ENLCN\cite{xia2022efficient} 26.18/0.8061}}
	\end{subfigure}
	\begin{subfigure}[]{0.15\textwidth}
		\centering
		\includegraphics[width=\textwidth]{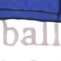}
		\caption*{\parbox[c]{15mm}{\textbf{HSPAN(Ours) 33.24/0.9351}}}
	\end{subfigure}

\vspace{4mm}
\begin{subfigure}[]{0.15\textwidth}
		\centering
		\includegraphics[width=\textwidth]{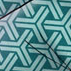}
		\caption*{\parbox[c]{8mm}{HR PSNR/SSIM}}
	\end{subfigure}
	\begin{subfigure}[]{0.15\textwidth}
		\centering
		\includegraphics[width=\textwidth]{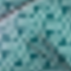}
		\caption*{\parbox[c]{8mm}{Bicubic 17.84/0.2758}}
	\end{subfigure}
	\begin{subfigure}[]{0.15\textwidth}
		\centering
		\includegraphics[width=\textwidth]{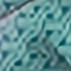}
		\caption*{\parbox[c]{15mm}{VDSR\cite{kim2016accurate} 18.21/0.3935}}
	\end{subfigure}
	\begin{subfigure}[]{0.15\textwidth}
		\centering
		\includegraphics[width=\textwidth]{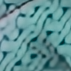}
		\caption*{\parbox[c]{18mm}{LapSRN\cite{lai2018fast} 17.73/0.3519}}
	\end{subfigure}
	\begin{subfigure}[]{0.15\textwidth}
		\centering
		\includegraphics[width=\textwidth]{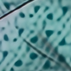}
		\caption*{\parbox[c]{15mm}{EDSR\cite{lim2017enhanced} 18.26/0.4244}}
	\end{subfigure}
	
\vspace{2mm}
	\begin{subfigure}[]{0.15\textwidth}
		\centering
		\includegraphics[width=\textwidth]{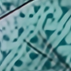}
		\caption*{\parbox[c]{16mm}{RCAN\cite{zhang2018image} 18.21/0.4586}}
	\end{subfigure}
	\begin{subfigure}[]{0.15\textwidth}
		\centering
		\includegraphics[width=\textwidth]{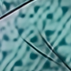}
		\caption*{\parbox[c]{15mm}{NLSN\cite{mei2021image} 18.51/0.4849}}
	\end{subfigure}
	\begin{subfigure}[]{0.15\textwidth}
		\centering
		\includegraphics[width=\textwidth]{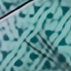}
		\caption*{\parbox[c]{16mm}{SwinIR\cite{liang2021swinir} 18.49/0.4902}}
	\end{subfigure}
	\begin{subfigure}[]{0.15\textwidth}
		\centering
		\includegraphics[width=\textwidth]{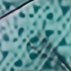}
		\caption*{\parbox[c]{17mm}{ENLCN\cite{xia2022efficient} 18.37/0.4423}}
	\end{subfigure}
	\begin{subfigure}[]{0.15\textwidth}
		\centering
		\includegraphics[width=\textwidth]{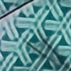}
		\caption*{\parbox[c]{15mm}{\textbf{HSPAN(Ours) 20.61/0.7000}}}
	\end{subfigure}
\end{minipage}
\hspace{-6mm}
\centering
   \caption{Visual comparisons on Urban100\cite{huang2015single} and Manga109\cite{matsui2017sketch} with scale factor 4.}
   \label{fig:visual-comparison-u100-m109}
\end{figure*}

\begin{table*}[!htbp]
\centering
\caption{Quantitative results on benchmark datasets with blur-downscale degradation. Best and second best results are \textbf{highlighted} and \underline{underlined}.}
\label{tab_x3_blur_psnr_ssim}
\begin{tabular}{|c|c|cc|cc|cc|cc|cc|}
\hline
\multirow{2}{*}{Method} &
  \multirow{2}{*}{Scale} &
  \multicolumn{2}{c|}{Set5\cite{bevilacqua2012low}} &
  \multicolumn{2}{c|}{Set14\cite{zeyde2010single}} &
  \multicolumn{2}{c|}{B100\cite{martin2001database}} &
  \multicolumn{2}{c|}{Urban100\cite{huang2015single}} &
  \multicolumn{2}{c|}{Manga109\cite{matsui2017sketch}} \\ \cline{3-12} 
 &
   &
  \multicolumn{1}{c|}{PSNR} &
  SSIM &
  \multicolumn{1}{c|}{PSNR} &
  SSIM &
  \multicolumn{1}{c|}{PSNR} &
  SSIM &
  \multicolumn{1}{c|}{PSNR} &
  SSIM &
  \multicolumn{1}{c|}{PSNR} &
  SSIM \\ \hline
\begin{tabular}[c]{@{}c@{}}Bicubic\\ SPMSR\cite{peleg2014statistical}\\ SRCNN\cite{dong2014learning}\\ FSRCNN\cite{dong2016accelerating}\\ VDSR\cite{kim2016accurate}\\ IRCNN\cite{zhang2017learning}\\ SRMDNF\cite{zhang2018learning}\\ RDN\cite{zhang2018residual}\\ EDSR\cite{lim2017enhanced}\end{tabular} &
  \begin{tabular}[c]{@{}c@{}}$\times 3$\\ $\times 3$\\ $\times 3$\\ $\times 3$\\ $\times 3$\\ $\times 3$\\ $\times 3$\\ $\times 3$\\ $\times 3$\end{tabular} &
  \multicolumn{1}{c|}{\begin{tabular}[c]{@{}c@{}}28.78\\ 32.21\\ 32.05\\ 32.33\\ 33.25\\ 33.38\\ 34.01\\ 34.58\\ 34.64\end{tabular}} &
  \begin{tabular}[c]{@{}c@{}}0.8308\\ 0.9001\\ 0.8944\\ 0.9020\\ 0.9150\\ 0.9182\\ 0.9242\\ 0.9280\\ 0.9282\end{tabular} &
  \multicolumn{1}{c|}{\begin{tabular}[c]{@{}c@{}}26.38\\ 28.89\\ 28.80\\ 28.91\\ 29.46\\ 29.63\\ 30.11\\ 30.53\\ 30.54\end{tabular}} &
  \begin{tabular}[c]{@{}c@{}}0.7271\\ 0.8105\\ 0.8074\\ 0.8122\\ 0.8244\\ 0.8281\\ 0.8364\\ 0.8447\\ 0.8451\end{tabular} &
  \multicolumn{1}{c|}{\begin{tabular}[c]{@{}c@{}}26.33\\ 28.13\\ 28.13\\ 28.17\\ 28.57\\ 28.65\\ 28.98\\ 29.23\\ 29.27\end{tabular}} &
  \begin{tabular}[c]{@{}c@{}}0.6918\\ 0.7740\\ 0.7736\\ 0.7791\\ 0.7893\\ 0.7922\\ 0.8009\\ 0.8079\\ 0.8094\end{tabular} &
  \multicolumn{1}{c|}{\begin{tabular}[c]{@{}c@{}}23.52\\ 25.84\\ 25.70\\ 25.71\\ 26.61\\ 26.77\\ 27.50\\ 28.46\\ 28.64\end{tabular}} &
  \begin{tabular}[c]{@{}c@{}}0.6862\\ 0.7856\\ 0.7770\\ 0.7842\\ 0.8136\\ 0.8154\\ 0.8370\\ 0.8582\\ 0.8618\end{tabular} &
  \multicolumn{1}{c|}{\begin{tabular}[c]{@{}c@{}}25.46\\ 29.64\\ 29.47\\ 29.37\\ 31.06\\ 31.15\\ 32.97\\ 33.97\\ 34.13\end{tabular}} &
  \begin{tabular}[c]{@{}c@{}}0.8149\\ 0.9003\\ 0.8924\\ 0.8985\\ 0.9234\\ 0.9245\\ 0.9391\\ 0.9465\\ 0.9477\end{tabular} \\
\begin{tabular}[c]{@{}c@{}}RCAN\cite{zhang2018image}\\ SAN\cite{dai2019second}\\ HAN\cite{niu2020single}\\ HSPAN (Ours)\\ HSPAN+ (Ours)\end{tabular} &
  \begin{tabular}[c]{@{}c@{}}$\times 3$\\ $\times 3$\\ $\times 3$\\ $\times 3$\\ $\times 3$\end{tabular} &
  \multicolumn{1}{c|}{\begin{tabular}[c]{@{}c@{}}34.70\\ 34.75\\ 34.76\\ \underline{34.90}\\ \textbf{35.00}\end{tabular}} &
  \begin{tabular}[c]{@{}c@{}}0.9288\\ 0.9290\\ 0.9294\\ \underline{0.9303}\\ \textbf{0.9310}\end{tabular} &
  \multicolumn{1}{c|}{\begin{tabular}[c]{@{}c@{}}30.63\\ 30.68\\ 30.70\\ \underline{30.81}\\ \textbf{30.91}\end{tabular}} &
  \begin{tabular}[c]{@{}c@{}}0.8462\\ 0.8466\\ 0.8475\\ \underline{0.8496}\\ \textbf{0.8508}\end{tabular} &
  \multicolumn{1}{c|}{\begin{tabular}[c]{@{}c@{}}29.32\\ 29.33\\ 29.34\\ \underline{29.42}\\ \textbf{29.48}\end{tabular}} &
  \begin{tabular}[c]{@{}c@{}}0.8093\\ 0.8101\\ 0.8106 \\ \underline{0.8131}\\ \textbf{0.8142}\end{tabular} &
  \multicolumn{1}{c|}{\begin{tabular}[c]{@{}c@{}}28.81\\ 28.83\\ 28.99\\ \underline{29.48}\\ \textbf{29.72}\end{tabular}} &
  \begin{tabular}[c]{@{}c@{}}0.8647\\ 0.8646\\ 0.8676\\ \underline{0.8766}\\ \textbf{0.8795}\end{tabular} &
  \multicolumn{1}{c|}{\begin{tabular}[c]{@{}c@{}}34.38\\ 34.46\\ 34.56\\ \underline{34.87} \\ \textbf{35.21}\end{tabular}} &
  \begin{tabular}[c]{@{}c@{}}0.9483\\ 0.9487\\ 0.9494\\ \underline{0.9515}\\ \textbf{0.9530}\end{tabular} \\ \hline
\end{tabular}
\end{table*}

\begin{figure*}[!htbp]
\centering
\hspace{6mm}
\begin{minipage}[]{0.50\textwidth}
	\begin{subfigure}[]{0.35\textwidth}
	\centering
	\includegraphics[width=1.7\textwidth,height=1.48\textwidth]{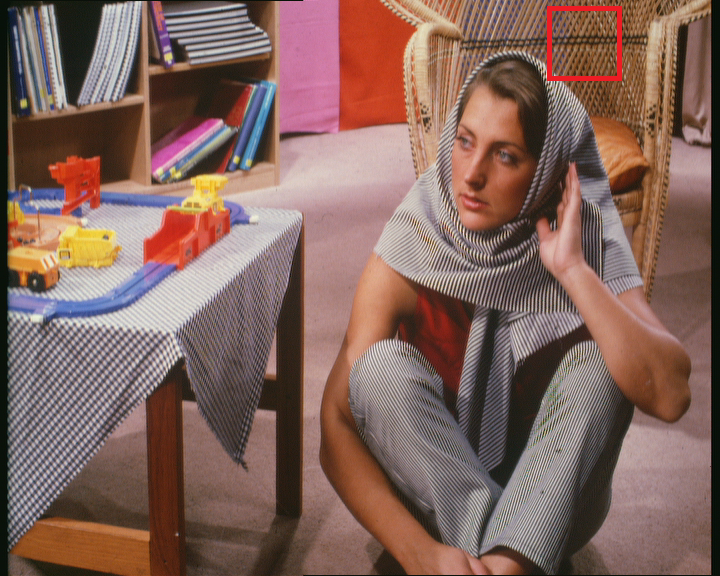}
	\end{subfigure}
	\vspace{3mm}

	\begin{subfigure}[]{0.186\textwidth}
	\centering
	\includegraphics[width=\textwidth]{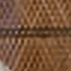}
	\captionsetup{font={scriptsize}}
	\caption*{\parbox[c]{14mm}{FSRCNN\cite{dong2016accelerating} 26.29/0.6845}}
	\end{subfigure}	
	\hspace{-0.8mm}
	\begin{subfigure}[]{0.186\textwidth}
	\centering
	\includegraphics[width=\textwidth]{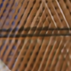}%
	\captionsetup{font={scriptsize}}
	\caption*{\parbox[c]{12mm}{\ EDSR\cite{lim2017enhanced} 26.81/0.7974}}
	\end{subfigure}
	\hspace{-0.8mm}
	\begin{subfigure}[]{0.186\textwidth}
	\centering
	\includegraphics[width=\textwidth]{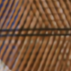}
	\captionsetup{font={scriptsize}}	
	\caption*{\parbox[c]{12mm}{\ \ HAN\cite{niu2020single} 26.43/0.7744}}
	\end{subfigure}	
\end{minipage}
\hspace{-36.5mm}
\begin{minipage}[]{0.15\textwidth}
	\vspace{-0.4mm}
	\begin{subfigure}[]{0.62\textwidth}
	\centering
	\includegraphics[width=\textwidth]{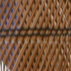}
	\captionsetup{font={scriptsize}}
	\caption*{HR \protect\\ PSNR/SSIM}
	\end{subfigure}

	\begin{subfigure}[]{0.62\textwidth}
	\centering
	\includegraphics[width=\textwidth]{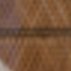}
	\captionsetup{font={scriptsize}}
	\caption*{\parbox[c]{10.5mm}{\ Bicubic 24.25/0.3970}}
	\vspace{2mm}
	\end{subfigure}

	\begin{subfigure}[]{0.62\textwidth}
	\centering
	\includegraphics[width=\textwidth]{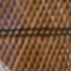}
	\captionsetup{font={scriptsize}}	
	\caption*{\parbox[c]{12mm}{\textbf{HSPAN(Ours)} \textbf{29.01}/\textbf{0.8731}}}
	\end{subfigure}
\end{minipage}
\begin{minipage}[]{0.50\textwidth}
	\begin{subfigure}[]{0.35\textwidth}
	\centering
	\includegraphics[width=1.7\textwidth,height=1.48\textwidth]{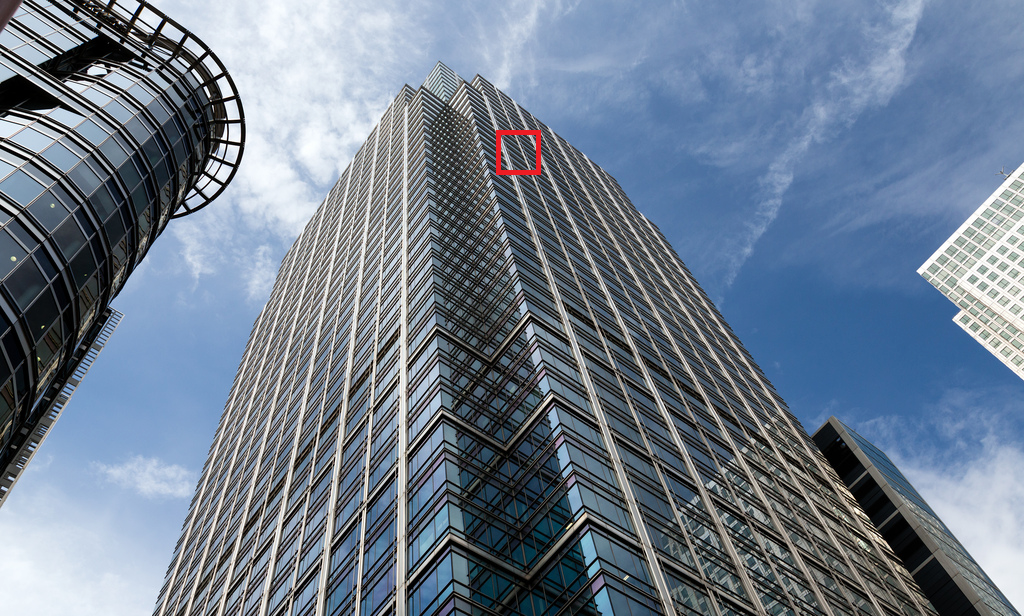}
	\end{subfigure}
	\vspace{3mm}

	\begin{subfigure}[]{0.185\textwidth}
	\centering
	\includegraphics[width=\textwidth]{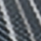}
	\captionsetup{font={scriptsize}}
	\caption*{\parbox[c]{14mm}{FSRCNN\cite{dong2016accelerating} 18.41/0.4069}}
	\end{subfigure}	
	\hspace{-0.8mm}
	\begin{subfigure}[]{0.185\textwidth}
	\centering
	\includegraphics[width=\textwidth]{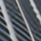}%
	\captionsetup{font={scriptsize}}
	\caption*{\parbox[c]{12mm}{\ EDSR\cite{lim2017enhanced} 19.18/0.5841}}
	\end{subfigure}
	\hspace{-0.8mm}
	\begin{subfigure}[]{0.185\textwidth}
	\centering
	\includegraphics[width=\textwidth]{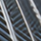}
	\captionsetup{font={scriptsize}}	
	\caption*{\parbox[c]{12mm}{\ \ HAN\cite{niu2020single} 18.01/0.4003}}
	\end{subfigure}	
\end{minipage}
\hspace{-36.5mm}
\begin{minipage}[]{0.15\textwidth}
	\vspace{-0.5mm}
	\begin{subfigure}[]{0.62\textwidth}
	\centering
	\includegraphics[width=\textwidth]{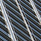}
	\captionsetup{font={scriptsize}}
	\caption*{HR \protect\\ PSNR/SSIM}
	\end{subfigure}

	\begin{subfigure}[]{0.62\textwidth}
	\centering
	\includegraphics[width=\textwidth]{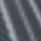}
	\captionsetup{font={scriptsize}}
	\caption*{\parbox[c]{10.5mm}{\ Bicubic 17.53/0.2550}}
	\vspace{2.1mm}
	\end{subfigure}

	\begin{subfigure}[]{0.62\textwidth}
	\centering
	\includegraphics[width=\textwidth]{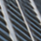}
	\captionsetup{font={scriptsize}}	
	\caption*{\parbox[c]{12mm}{\textbf{HSPAN(Ours)} \textbf{20.64}/\textbf{0.7618}}}
	\end{subfigure}
\end{minipage}
\hspace{-4mm}
\centering
   \caption{Visual comparisons for $\times$3 SISR with blur-downscale degradation on the image 'barbara' and 'img\_047' from Set14\cite{zeyde2010single} and Urban100\cite{huang2015single}, respectively. Best and second best results are \textbf{highlighted} and \underline{underlined}. Please zoom in for best view.}
   \label{fig:visual-comparison-blur-set14}
\end{figure*}

\subsection{Comparisons with State-of-the-art}
\subsubsection{Bicubic-downscale degradation}
\ \par
We compare our HSPAN with 18 state-of-the-art methods including FSRCNN\cite{dong2016accelerating}, VDSR\cite{kim2016accurate}, LapSRN\cite{lai2018fast}, EDSR\cite{lim2017enhanced}, MemNet\cite{tai2017memnet}, SRMDNF\cite{zhang2018learning}, DBPN\cite{haris2018deep}, RDN\cite{zhang2018residual}, RCAN\cite{zhang2018image}, SAN\cite{dai2019second}, OISR\cite{he2019ode},IGNN\cite{zhou2020cross}, CSNLN\cite{mei2020image}, HAN\cite{niu2020single}, NLSN\cite{mei2021image}, DRLN\cite{anwar2022densely}, SwinIR\cite{liang2021swinir}, and ENLCN\cite{xia2022efficient}. HSPAN+ is the self ensemble results of our HSPAN.
\vspace{1mm}

\noindent\textbf{Quantitative Evaluations.} The quantitative comparisons on five benchmark datasets with different scale factors $\times 2$, $\times 3$ and $\times 4$ are shown in \cref{tab:x2_x3_x4_psnr_ssim}, from which we can see that our HSPAN outperforms other state-of-the-art models by a large margin on almost all scale factors and benchmarks. For example, compared with the recent state-of-the-art NLSN\cite{mei2021image} in scale factor $\times 4$, our HSPAN brings 0.23dB, 0.16dB, 0.08dB, 0.40dB and 0.47dB improvement on Set5, Set14, B100, Urban100 and Manga109 datasets, respectively. It is worth mentioning that the HSPAN is designed to utilize self-similarity information. Thus, the proposed HSPAN can achieve impressive reconstruction results especially for more challenging datasets Urban100 and Manga109, which contain a large amount of self-similarity information. 

On Urban100 dataset ($\times 3$), which is specially designed to test self-similarity-based SISR methods, we can see that recent deep SISR methods have very limited improvements on this dataset. For example, compared with CSNLN \cite{mei2020image}, NLSN\cite{mei2021image} brings 0.12dB improvement. In contrast, our HSPAN brings 0.54dB improvement on this dataset compared to CSNLN. The impressive improvement obtained by the HSPAN on Urban100 dataset is consistent with our motivation to design the HSAP, which aims to explore the self-similarity information efficiently.

%\vspace{1mm}
%%Our DLSN is constructed by inserting the five proposed SSL modules into the EDSR backbone, which implies that the self-similarity information explored by the proposed SSL module can significantly improve the reconstruction performance of EDSR and even makes the improvd EDSR outperform very impressive RCAN\cite{zhang2018image}, SAN\cite{dai2019second} and NLSN\cite{mei2021image} etc. 

\noindent\textbf{Qualitative Evaluations.} Visual comparisons on the challenging datasets Urban100 and Mange109 ($\times 4$) are shown in \cref{fig:visual-comparison-u100-m109}, from which we can see that the proposed HSPAN can repair the severely damaged textures when there are informative self-similarity textures in the input LR image. By comparing the reconstructed textures of image 'img\_045' in \cref{fig:visual-comparison-u100-m109}, we can observe that the generated textures of our HSPAN are similar to the HR textures, but other very competitive deep SISR models without non-local attention, such as RDN\cite{zhang2018residual} and RCAN\cite{zhang2018image}, cannot repair such severely damaged regions. Furthermore, compared with other deep SISR models based on non-local attention such as SwinIR\cite{liang2021swinir} and ENLCN\cite{xia2022efficient}, our HSPAN still achieves better reconstruction performance with more accurate image details. These comparisons indicate that our HSPAN is more efficient in repairing severely damaged regions by fusing the self-similarity information with the proposed HSPA. 

By comparing the generated textures of image 'GOODKISSVer2' in \cref{fig:visual-comparison-u100-m109}, we can see that our HSPAN is the only method that can accurately restore the texture of the word "ball" in HR, and the restoration results of other methods are inconsistent with the textures in HR. Furthermore, in image 'GOODKISSVer2', our HSPAN outperforms the competitive SwinIR by 2.29dB. These visual comparisons demonstrate that our HSPAN not only outperforms other deep SISR models in quantitative metrics, but also significantly improves reconstructed textures perceptually.

\subsubsection{Blur-downscale degradation}
\ \par
In the blur-downscale degradation SISR tasks, the SR performance is verified at scale factor $\times 3$ and the gaussian standard deviation is set to 1.6 as previous works\cite{zhang2018learning,zhang2018image}. Our HSPAN are compared with 11 state-of-the-art methods: SPMSR\cite{peleg2014statistical}, SRCNN\cite{dong2014learning}, FSRCNN\cite{dong2016accelerating}, VDSR\cite{kim2016accurate}, IRCNN\cite{zhang2017learning}, EDSR\cite{lim2017enhanced}, SRMDNF\cite{zhang2018learning}, RDN\cite{zhang2018residual}, RCAN\cite{zhang2018image}, SAN\cite{dai2019second} and HAN\cite{niu2020single}.

\noindent\textbf{Quantitative Evaluations.} 
We compare the quantitative results with the blur-downscale degradation at scale factor $\times 3$. As shown in \cref{tab_x3_blur_psnr_ssim}, we can see that the proposed HSPAN achieves best results on all benchmark datasets. On natural image datasets Set5, Set14, and B100, our HSPAN achieves a satisfactory reconstruction performance. Furthermore, our HSPAN can still bring a large performance improvement on the challenging datasets Urban100 and Manga109. In particular, compared with the competitive HAN\cite{niu2020single}, our HSPAN brings 0.49dB and 0.31dB improvement on Urban100 and Manga109 datasets, respectively. This is consistent with previous experiment on bicubic-downscale degradation. These results indicate that our HSPAN is not limited to the degradation type and can still achieve very impressive SR performance on the SISR task with blur-downscale degradation.

%\begin{table}[!htbp]
%\centering
%\caption{Quantitative comparisons on Set5\cite{bevilacqua2012low} ($\times 3$) with noise-downscale degradation in $5\times10^4$ iterations. Best and second best results are \textbf{highlighted} and \underline{underlined}.}
%\label{tab_x3_noise_psnr_ssim}
%\resizebox{0.49\textwidth}{!}{
%\begin{tabular}{|ccccc|}
%\hline
%\multirow{2}{*}{Noise level} & FSRCNN\cite{dong2016accelerating}       & EDSR\cite{lim2017enhanced}        & RCAN\cite{zhang2018image}          & HSPAN(Ours)            \\ \cline{2-5} 
%                             & PSNR/SSIM    & PSNR/SSIM    & PSNR/SSIM    & PSNR/SSIM             \\ \hline
%10                           & 29.80/0.8386 & \underline{31.73}/\underline{0.8795} & 31.70/0.8790 & \textbf{31.85}/\textbf{0.8803} \\
%15                           & 28.80/0.8119 & \underline{30.68}/\underline{0.8594} & 30.65/0.8579 & \textbf{30.80}/\textbf{0.8607} \\
%20                           & 27.97/0.7893 & \underline{29.82}/\underline{0.8416} & 29.79/0.8399 & \textbf{29.92}/\textbf{0.8417} \\
%25                           & 27.31/0.7719 & \underline{29.08}/\underline{0.8242} & \underline{29.08}/0.8231 & \textbf{29.20}/\textbf{0.8269} \\ \hline
%\end{tabular}
%}
%\end{table}

\noindent\textbf{Qualitative Evaluations.} 
Visual comparisons on datasets Set5 and Urban100 with blur-downscale degradation are shown in \cref{fig:visual-comparison-blur-set14}, from which we can see that our HSPAN reconstructs the most visual pleasing results with more accurate textures. By observing the results of Bicubic interpolation we can find that the textures in the input LR image has been completely damaged. Thus, it is impossible to accurately recover the desired textures with only local information. This means that the reconstruction results of each method presented in \cref{fig:visual-comparison-blur-set14} can be used to evaluate the ability of each method in exploring self-similarity information. The reconstructed textures of our HSPAN is very close to the HR image, while the competitive HAN\cite{niu2020single} cannot restore textures that are severely damaged by blur-downscale degradation. For example, in image 'img\_047' from \cref{fig:visual-comparison-blur-set14}, HAN cannot generate satisfactory textures, even though the selected region has valuable repeated structured architectural textures in the input image. These visual comparisons demonstrate that our HSPAN is indeed effective in the blur-downscale SISR task.

%\subsubsection{Noisy-downscale degradation}
%\ \par
%Since most state-of-the-art methods do not provide the verification of SR performance under noisy-downscale degradation, we reimplemented some deep SISR models including FSRCNN, EDSR and RCAN under the same noisy-downscale degradation. The SR performance of these state-of-the-art SISR methods at noise levels of 10, 15, 20, and 25 are shown in \cref{tab_x3_noise_psnr_ssim}, from which we can see that our HSPAN outperforms other state-of-the-art SISR models at all noise levels, which indicates that our HSPA is robust in handling SISR tasks with different noise levels.

\section{Conclusions}
In this paper, we provided new insights into the NLA used in SISR problems and found that the softmax transformation, a key component of the NLA, is not suitable for exploring long-range information. To overcome this drawback, we designed a flexible high-similarity-pass attention (HSPA) that enables our deep high-similarity-pass attention network (HSPAN) to focus on more valuable non-local textures while removing irrelevant ones. Furthermore, we explored some key properties of the proposed soft thresholding (ST) operation to train our HSPA in an end-to-end manner. To the best of our knowledge, this is the first attempt to analyze and address the limitations of utilizing the softmax transformation for long-range sequence modeling in low-level vision problems. In addition, extensive experiments demonstrate that our HSPA and ST operation can be integrated as efficient general building units in existing deep SISR models.

\bibliographystyle{IEEEtranS}
\bibliography{main}
\end{document}